\def\SkillHorizonArxiv{1}
\documentclass{article}
\usepackage{iclr2027_conference,times}
\usepackage[T1]{fontenc}
\usepackage[utf8]{inputenc}
\usepackage{amsmath,amssymb,booktabs,tabularx,array,graphicx,xcolor}
\usepackage{colortbl}
\usepackage{microtype,placeins,wrapfig}
\usepackage{needspace}
\usepackage{fvextra}
\DefineVerbatimEnvironment{PaperPrompt}{Verbatim}{fontsize=\footnotesize,
  breaklines=true,breakanywhere=true,samepage=true,frame=single,framesep=5pt}
\usepackage{tikz}
\usetikzlibrary{arrows.meta,positioning,calc}
\usepackage[hyphens]{url}
\usepackage[hidelinks]{hyperref}
\newcolumntype{Y}{>{\raggedright\arraybackslash}X}
\newcommand{\bench}{SkillHorizon}
\newcommand{\skillfile}{\texttt{SKILL.md}}

\newcommand{\scorepair}[2]{#1\,/\,#2}

\title{SkillHorizon: Constructing Long-Horizon Agent Tasks through Skill Composition}
\author{Anonymous authors}
\hypersetup{pdftitle={SkillHorizon: Constructing Long-Horizon Agent Tasks through Skill Composition},pdfauthor={}}
\ifdefined\SkillHorizonArxiv
  \iclrfinalcopy
\author{%
\begin{minipage}{\dimexpr\textwidth-2\tabcolsep\relax}
\raggedright\normalfont\fontsize{10}{12}\selectfont
Yuxuan Liu\textsuperscript{1,*},
Zhaochen Su\textsuperscript{1,*},
Hangyu Guo\textsuperscript{2},
Shijue Huang\textsuperscript{1},
Zhixiong Zhang\textsuperscript{3},\\
Kawai Chung\textsuperscript{1},
Zhongwei Xie\textsuperscript{1},
Tianshi Zheng\textsuperscript{1},
Wei Fan\textsuperscript{1},
Qing Zong\textsuperscript{1},\\
Haochen Shi\textsuperscript{1},
Baixuan Xu\textsuperscript{1},
Haoran Li\textsuperscript{1,$\dagger$},
Yangqiu Song\textsuperscript{1}\\[3pt]
\textsuperscript{1}The Hong Kong University of Science and Technology\\
\textsuperscript{2}Alibaba Group\\
\textsuperscript{3}Shanghai Jiao Tong University\\
{\fontsize{9}{11}\selectfont
\texttt{\{\href{mailto:yliurk@connect.ust.hk}{yliurk},\href{mailto:zsubf@connect.ust.hk}{zsubf},\href{mailto:hlibt@connect.ust.hk}{hlibt}\}@connect.ust.hk},
\href{mailto:yqsong@cse.ust.hk}{\texttt{yqsong@cse.ust.hk}}\\[3pt]
\textsuperscript{*}Equal contribution.\quad
\textsuperscript{$\dagger$}Corresponding author.}
\end{minipage}}
\hypersetup{pdfauthor={Yuxuan Liu, Zhaochen Su, Hangyu Guo, Shijue Huang, Zhixiong Zhang, Kawai Chung, Zhongwei Xie, Tianshi Zheng, Wei Fan, Qing Zong, Haochen Shi, Baixuan Xu, Haoran Li, Yangqiu Song}}

\fi
\begin{document}
\raggedbottom
\maketitle
\ifdefined\SkillHorizonArxiv
  \lhead{Preprint}
\fi
\begingroup
\clubpenalty=10000
\widowpenalty=10000
\displaywidowpenalty=10000
\begin{abstract}
Long-horizon agents must complete workflows in which later activities
depend on earlier outputs and changes to shared state. A key challenge
for evaluation is to expand task requirements and their dependencies
while keeping tasks executable and outcomes verifiable. We introduce
\bench{}, a benchmark that constructs long-horizon tasks through skill
composition in seven business environments. Its generators use
domain-specific rules to turn reusable skill procedures into concrete
business activities. These activities are connected through input--output
dependencies, shared resources, and state updates, allowing workload and
coordination demands to grow together. Successful reference executions
validate task feasibility and measure interaction workload, while
programmatic evaluators assess local completion and whole-task success.
We evaluate five language models on 1,050 tasks with reference interaction
workloads ranging from 64K to 1M tokens. Across these scales, mean
local-goal completion falls from 90.4\% to 46.8\%, as completed work fails
to keep pace with growing requirements. At larger workloads, agents more
often submit unsuccessful final answers with substantial work unfinished
and tool-call allowance remaining. Trace analysis also identifies cases
where scripts execute successfully but their outputs fail downstream
business checks. \bench{} provides a configurable testbed for studying
sustained progress, dependency coordination, and complete task delivery
in growing workflows. Our code is available at
\url{https://github.com/KnowComp/SkillHorizon}.
\end{abstract}
\suppressfloats[t]
\section{Introduction}
\label{sec:introduction}

\begin{figure}[t]
\centering
\includegraphics[width=\linewidth,height=.37797\linewidth]{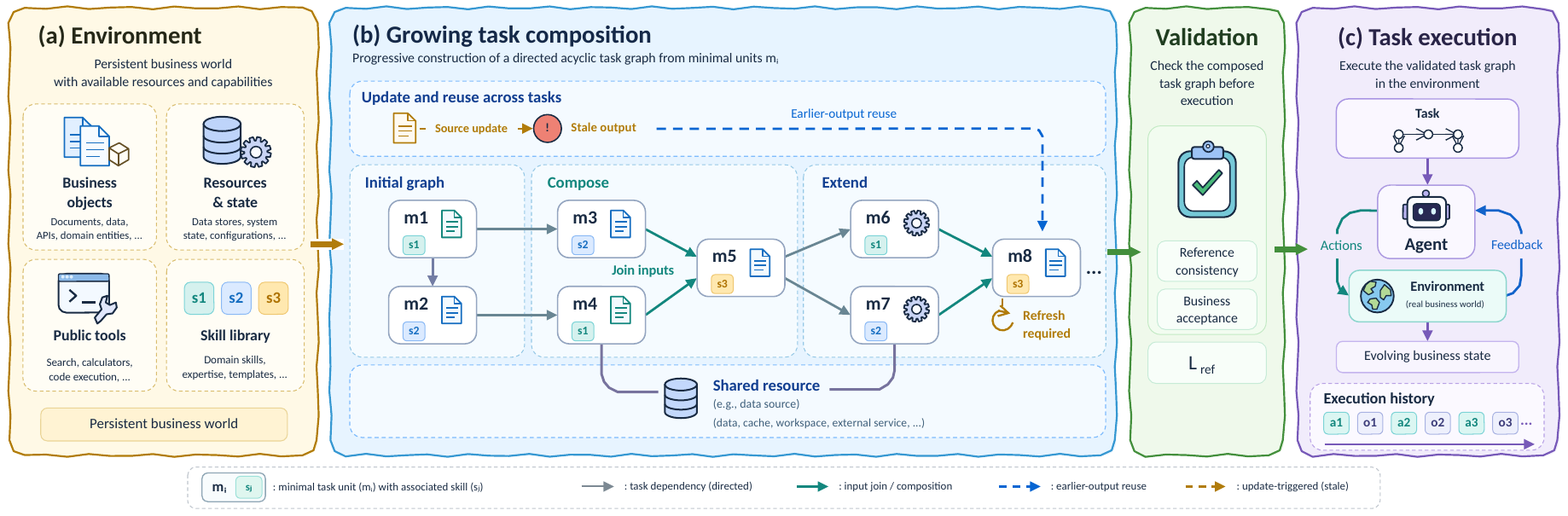}
\caption{SkillHorizon constructs long-horizon tasks through skill composition. (a) Environments provide objects, resources, tools, and skills. (b) Tasks grow through dependencies, resource constraints, and updates requiring downstream refresh. Public reference executions validate business completion and measure $L_{\mathrm{ref}}$. (c) Agents execute validated tasks through repeated interactions, changing business state and accumulating actions $a_i$ and observations $o_i$.}
\label{fig:construction}
\end{figure}

Language agents increasingly carry out business workflows that involve
multiple activities and sustained interaction with tools and
data~\citep{xu2024theagentcompany,li2025toolathlon}. The outputs of one
activity become inputs to later stages, while resource commitments shape
subsequent choices. Changes in the environment may also require completed
work to be reviewed or revised. Agents must therefore maintain valid
intermediate results and coordinate activities as they work toward the
final goal. Studying this ability requires executable tasks that can grow
in required work and dependencies, with persistent state and verifiable
outcomes.

Existing research provides foundations for long-horizon evaluation and
skill reuse. WorkArena++ combines atomic tasks into enterprise workflows
with reference executions and validation
functions~\citep{boisvert2024workarena}. TheAgentCompany evaluates work
across business applications, code execution, and communication in a
simulated workplace~\citep{xu2024theagentcompany}, while AgentLongBench
studies information use over extended interaction
histories~\citep{fang2026agentlongbench}. In parallel, Voyager develops an
executable skill library, and SkillsBench evaluates packaged skills
across tasks~\citep{wang2023voyager,li2026skillsbench}. AgentSkillOS further
studies skill retrieval and graph-based
coordination~\citep{li2026agentskillos}. These studies support complex
workflows and procedural reuse.

Despite this progress, existing evaluations offer limited insight into
how agents perform as more business activities are added and linked to
earlier work. Studying this scaling requires a construction method that
expands task goals and dependencies together. Each new activity must
specify its required inputs, expected outputs, and use of shared
resources. State updates must also define when earlier results need to
be revised. At the same time, resource allocations and completion checks
must match the expanded task. The key challenge is to build these larger
workflows from reusable skills while keeping them executable and their
outcomes verifiable.

We introduce \bench{}, a benchmark that constructs long-horizon tasks
through skill composition in seven business environments spanning finance,
field maintenance, social campaigns, logistics, research, documentation,
and manufacturing. Domain-specific generators turn reusable skill procedures
into business modules with concrete objects and goals. They connect modules
through input--output dependencies and configure shared resources and state
updates. Adding modules increases required work, while their connections
determine which earlier results must be reused, combined, or revised. The
generators create task goals and initial states together. Each environment
provides public tools, skill resources, and programmatic evaluators.
Successful reference executions validate task feasibility and measure
interaction workload; evaluators check local-goal completion and whole-task
acceptance. Figure~\ref{fig:construction} summarizes the pipeline.

We evaluate five language models on 1,050 tasks with reference interaction
workloads ranging from 64K to 1M tokens, yielding 5,250 model--task
executions. Mean local-goal completion falls from 90.4\% to 46.8\% across
these scales, as completed work fails to keep pace with growing
requirements. We also analyze failed final answers with recorded
completion and tool use. At 1M, 54.0\% of these runs complete less than
25\% of local goals while retaining over 50\% of the tool-call allowance.
The corresponding share is 2.4\% at 64K. Trace analysis further identifies
cases where scripts execute successfully but their outputs fail
downstream business checks. These findings reveal difficulties in
sustaining progress and using local results to complete larger workflows.

Our contributions are threefold:
\par\noindent\textbf{1. Skill-based task construction.}
We introduce a method that builds business activities from reusable
skills and connects them through input--output dependencies, shared
resources, and state updates. The method supports configurable expansion
of required work and coordination demands.
\par\noindent\textbf{2. An executable and scalable benchmark.}
We build seven persistent business environments with task generators,
public tools, skill resources, reference executions, and programmatic
evaluators. These resources support reproducible evaluation across task
instances and workload scales.
\par\noindent\textbf{3. An empirical study of long-horizon execution.}
We evaluate five models across increasing workloads and identify gaps
between required and completed work, including unfinished execution with
unused tool allowance and downstream failures of script outputs.

\section{The SkillHorizon Benchmark}
\label{sec:task}

\subsection{Business Environments and Skills}
\label{sec:environments}

SkillHorizon comprises seven executable business environments: FinanceOps, FieldOps, SocialOps, LogisticsOps, ResearchOps, DocOps, and FactoryOps. Each environment maintains a persistent business world of objects, resources, and artifacts, governed by domain-specific rules. Operations produce artifacts, allocate resources, and update business state. Their effects persist across interactions: later activities consume earlier outputs and must satisfy the resource and validity constraints of the current state. Appendix~\ref{app:environment-resources} details the resources and business rules of each environment.

Business operations are exposed as Model Context Protocol (MCP) tools \citep{mcp2025spec}. Through domain-specific gateways, agents query business objects, invoke operations, and inspect artifacts. The environment checks requested operations against the current state and applicable business constraints, returning execution results or diagnostic feedback.

\label{sec:skill-grounding}
Each environment $\mathcal E$ provides a library $\mathcal S_{\mathcal E}$ of reusable skills that organize public tool operations into business procedures. We characterize a skill $s\in\mathcal S_{\mathcal E}$ as
\begin{equation}
s=(\alpha_s,I_s,P_s,O_s),
\label{eq:skill}
\end{equation}
where $\alpha_s$ specifies applicability conditions, $I_s$ required inputs, $P_s$ the procedure, and $O_s$ expected outputs and state effects. Skills follow the Agent Skills package format \citep{agentskills_spec}, with procedural documentation and optional scripts and reference materials. During task construction, domain-specific rules instantiate these requirements and effects over concrete business objects to define local work. Agents select and apply these procedures through public tools.

\subsection{Dependency-Structured Task Construction}
\label{sec:construction}

A \emph{business module} is the unit of task composition. It specifies a piece of required business work by applying one or more skill procedures to concrete objects and inputs. Domain-specific rules determine its input requirements $I_m$, required outputs and state effects $O_m$, and completion conditions $g_m$:
\begin{equation}
m=(I_m,O_m,g_m).
\label{eq:business-module}
\end{equation}
A module may combine several skills to fulfill its business requirements, while the same skill can be reused across modules with different objects and goals.

We connect modules when completing one requires outputs produced by another. The resulting dependencies form a directed acyclic graph $G=(M,D)$. Each edge $e=(m_u,m_v,d_e)\in D$ specifies the outputs that $m_v$ requires from $m_u$, together with conditions on their content, provenance, quantity, or validity. Environment operations and completion checks enforce these requirements. Shared resources and mutable business state further constrain when modules can proceed and whether previously produced outputs remain usable.

For a construction configuration $c$ and sampling seed $\xi$, the environment-specific generator produces
\begin{equation}
(G,x_0,g)=\operatorname{Gen}_{\mathcal E}(c,\xi).
\label{eq:instance-generation}
\end{equation}
The configuration controls work scope, dependency patterns, resource conditions, and updates. The seed selects concrete objects, attributes, and connections within these rules. Construction couples the dependency graph $G$ with an initial state $x_0$ that supplies the corresponding business objects and resources, and a goal $g$ that requires the generated work and its joint delivery conditions to be satisfied. Agents pursue this goal through public environment operations.

We extend tasks by adding required business modules and connecting them to existing work through domain-specific composition rules. Sequential extension makes an earlier output a prerequisite for further work; joins require multiple upstream outputs to jointly satisfy a downstream requirement; long-range reuse carries earlier results into later stages. These connections determine which results must be combined and how far their content and validity must be maintained. Task expansion therefore controls both the amount of required work and the dependency structure through which that work contributes to the overall goal.

Task expansion also determines the business state needed to support the generated work. Domain-specific rules derive resource quantities and allocations from task requirements while imposing capacity and scheduling constraints. For tasks with updates, these rules specify when changes occur and which artifacts or subsequent operations they affect. Existing results retain or lose validity according to the domain rules, creating requirements for revision and downstream coordination. Figure~\ref{fig:construction} summarizes the construction pipeline, and Appendix~\ref{app:stage-construction} details the domain-specific rules. Public reference execution then verifies that each instance's business requirements can be satisfied.

\subsection{Reference Validation and Workload Calibration}
\label{sec:length}

We validate generated instances with a rule-based reference policy that executes business activities through public environment tools, following the executable-oracle approach used in compositional benchmarks \citep{boisvert2024workarena}. The policy queries the current state and carries out the required work under the instance's dependency, resource, and update conditions. An instance passes validation when its resulting artifacts and final business state satisfy the local completion requirements and global delivery constraints. We calibrate workload from these successful reference traces.

We measure reference workload as the cumulative tool-interaction volume of a successful reference execution. For a trace containing $R$ tool calls, we define
\begin{equation}
L_{\mathrm{ref}}=
\sum_{i=1}^{R}
\left[\operatorname{Tokens}(a_i)+\operatorname{Tokens}(o_i)\right],
\label{eq:reference-workload}
\end{equation}
where $a_i$ contains the tool name and arguments, and $o_i$ is the returned observation. Fixed serialization and tokenization count each call--observation pair, including repeated queries. Thus, $L_{\mathrm{ref}}$ measures tool interaction as the reference policy fulfills the task's business requirements.

To obtain tasks at different workload scales, we adjust domain-specific work-scale parameters under the selected construction rules and measure the resulting successful reference executions. Instances are assigned to workload bands according to their measured $L_{\mathrm{ref}}$. Alongside this measure, we record business work counts and dependency properties, including depth, joins, and reuse spans where applicable, to characterize the amount and organization of required work. Appendix~\ref{app:length} details the calibration parameters, workload-band boundaries, and instance-selection rules.

\section{Experiments}
\label{sec:experiments}

We evaluate agents on generated tasks that vary in the amount of required business work and the dependencies connecting it. Across seven executable environments, we measure terminal business-goal completion and whole-task success to examine agents' ability to sustain progress toward an overall goal.\looseness=-1

\subsection{Experimental Setup}
\paragraph{Models and harness.}
We evaluate Grok 4.6, Qwen3.8-Flash, DeepSeek V4.1 Flash, Gemini 3.8 Flash, and GPT-5.6 Sol using a unified Harbor + Inspect framework across all seven environments \citep{harbor,inspect2024}. Agents interact with persistent business state through public tools, guided by skill documents and supplied scripts. Summary-based context compaction is triggered when estimated input length reaches 200,000 tokens, within an input capacity of 480,000 tokens \citep{anthropic2025context}. Business state and generated artifacts remain accessible through public tools. Appendix~\ref{app:protocols} specifies model configurations and execution settings.

\paragraph{Task set.}
Tasks span seven business environments and five bands: 64K, 128K, 256K, 512K, and 1M tokens, assigned by measured reference workload $L_{\mathrm{ref}}$. Each band contains 210 tasks generated under approximately 100 parameter configurations spanning business scale, dependency structure, and update conditions. Appendix~\ref{app:length} details task sampling and construction profiles.

\paragraph{Tool budget.}
We scale tool-attempt allowances using each task's public reference execution:
\[
B_{\mathrm{ref}}=R+\lceil\lambda A\rceil,\qquad \lambda=0.5,
\]
where $R$ counts all reference tool calls and $A$ counts business-execution calls. Queries, business operations, and corrective attempts share the allowance. Appendix~\ref{app:protocols} specifies accounting and actual allowances.

\begin{samepage}
\paragraph{Evaluation metrics.}
\label{sec:evaluation}

We evaluate task outcomes by checking the final environment state \citep{trivedi2024appworld,yao2024taubench}. Local-goal completion (LGC) measures the fraction of local business goals satisfied at termination, averaged equally across tasks. Task success (TS) measures the proportion of tasks satisfying all final requirements, including cross-artifact dependencies and delivery constraints. Appendix~\ref{app:metrics} details the scoring and aggregation rules.
\par
\end{samepage}

\ifdefined\SkillHorizonArxiv\suppressfloats[t]\fi
\subsection{Main Results}
\begin{table}[!htbp]
\centering\small
\caption{Main results across seven environments and reference-workload bands (\%). Local-goal completion (LGC) averages terminal native-goal fractions; task success (TS) measures full business acceptance. Both metrics use the same 210 executions per model and band, balanced across environments. Overall averages all 1,050 executions per model.}
\label{tab:main-results}
\setlength{\tabcolsep}{3pt}
\begin{tabular*}{\linewidth}{@{\extracolsep{\fill}}llrrrrrr@{}}
\toprule
Model & Metric & 64K & 128K & 256K & 512K & 1M & Overall \\
\midrule
Grok 4.6 & LGC $\uparrow$ & 95.5 & 93.1 & 89.2 & 78.7 & 73.4 & 86.0 \\
 & TS $\uparrow$ & 90.0 & 81.4 & 74.8 & 60.5 & 52.4 & 71.8 \\
\addlinespace[3pt]
Qwen3.8-Flash & LGC $\uparrow$ & 94.3 & 83.1 & 73.1 & 61.6 & 48.9 & 72.2 \\
 & TS $\uparrow$ & 82.9 & 61.4 & 45.2 & 39.5 & 22.9 & 50.4 \\
\addlinespace[3pt]
DeepSeek V4.1 Flash & LGC $\uparrow$ & 88.9 & 79.1 & 72.5 & 63.9 & 56.1 & 72.1 \\
 & TS $\uparrow$ & 74.8 & 53.8 & 43.8 & 37.1 & 26.2 & 47.1 \\
\addlinespace[3pt]
Gemini 3.8 Flash & LGC $\uparrow$ & 92.3 & 81.3 & 64.1 & 38.5 & 21.3 & 59.5 \\
 & TS $\uparrow$ & 78.6 & 58.6 & 44.3 & 25.2 & 8.1 & 43.0 \\
\addlinespace[3pt]
GPT-5.6 Sol & LGC $\uparrow$ & 80.9 & 63.4 & 56.2 & 44.9 & 34.1 & 55.9 \\
 & TS $\uparrow$ & 51.4 & 23.3 & 10.5 & 5.7 & 1.9 & 18.6 \\
\bottomrule
\end{tabular*}
\end{table}

\begin{figure}[!htbp]
\centering
\includegraphics[width=\linewidth]{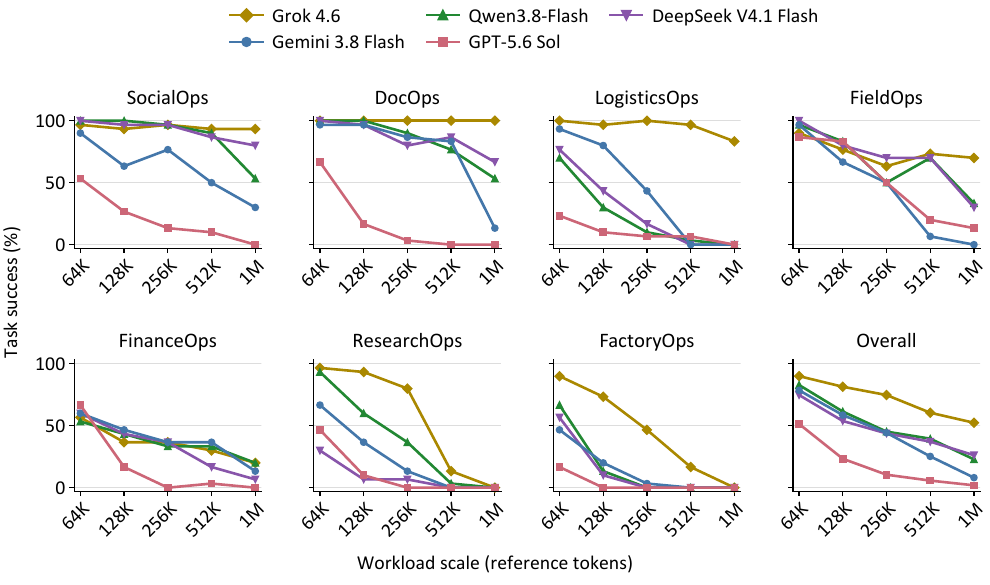}
\caption{Task success across seven business environments and reference-workload scales. The Overall panel pools all seven environments with equal task weight. Colors and markers identify models; each environment point includes 30 assigned tasks and each Overall point includes 210. Table~\ref{tab:detailed-results} reports task success and local-goal completion.}
\label{fig:environment-curves}
\end{figure}

\textbf{Performance declines with workload, and model rankings vary across scales.} As shown in Table~\ref{tab:main-results}, all five models achieve lower LGC and TS at 1M than at 64K. Grok leads overall with 86.0\% LGC and 71.8\% TS, but its LGC falls from 95.5\% to 73.4\%, and its TS from 90.0\% to 52.4\%. Qwen and DeepSeek have similar overall LGC, at 72.2\% and 72.1\%, respectively, yet Qwen leads by 5.4 percentage points at 64K while DeepSeek leads by 7.2 points at 1M. Their TS ranking also reverses. Overall averages therefore do not fully characterize performance on longer tasks.

\textbf{High local-goal completion can coexist with low task success.} From 64K to 1M, Gemini's FieldOps LGC falls from 98.6\% to 4.6\%, while TS declines from 96.7\% to zero. GPT's FinanceOps LGC remains high, declining from 93.7\% to 85.8\%, while TS falls from 66.7\% to zero.

\ifdefined\SkillHorizonArxiv\else\newpage\fi
\textbf{Task success follows different workload-scaling patterns across environments.} As shown in Figure~\ref{fig:environment-curves}, Grok, DeepSeek, and Qwen each retain task success above 50\% at 1M in SocialOps and DocOps. LogisticsOps and FieldOps show wider differences between models: Grok reaches 83.3\% and 70.0\%, respectively, while the other models achieve zero success in LogisticsOps and at most 33.3\% in FieldOps. In FinanceOps, no model exceeds 20.0\% task success at 1M. In ResearchOps and FactoryOps, all five models reach zero task success at this scale.

\FloatBarrier
\section{Analysis}
\label{sec:behavior}
We analyze execution traces to examine progress under growing workloads, the delivery of local results, recovery after failed attempts, and the use of historical outputs across compaction boundaries. We also examine how task sampling affects the measured success rate.

\subsection{Sustained Progress and Completion}
\begin{figure}[!htb]
\centering
\includegraphics[width=.90\linewidth]{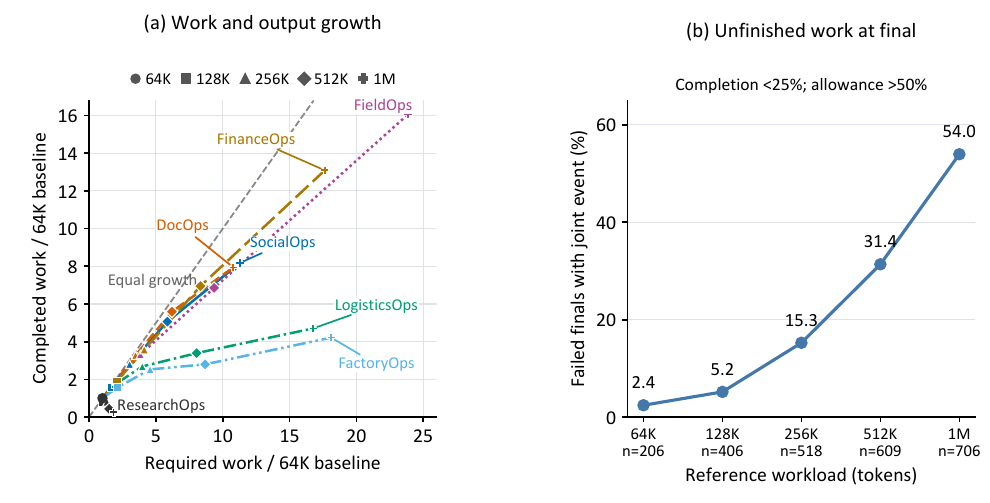}
\caption{Growing requirements outpace effective output. (a) Required and completed goal counts across five workload bands, averaged over five models and separately normalized to each environment's 64K baseline. The gray diagonal marks equal proportional growth. (b) Percentage of failed finals with both measurements that complete less than 25\% of goals while leaving more than 50\% of the tool allowance unused.}
\label{fig:sustained-progress}
\end{figure}

\textbf{Effective output does not keep pace with growing task requirements.} Figure~\ref{fig:sustained-progress}a traces the growth of required and completed work across the five workload bands. Averaged across models, the number of completed local goals increases from 64K to 1M in six environments, but grows more slowly than the number required. In ResearchOps, the average number of completed rounds decreases over this range even as task requirements grow. Mean LGC falls from 90.4\% to 46.8\%. The shortfall also becomes more widespread: the proportion of executions completing less than one-quarter of their local goals rises from 2.3\% to 47.1\%. Appendix~\ref{app:sustained-progress} reports absolute counts by model.

\textbf{Agents increasingly stop with substantial work and tool allowance remaining.} Among failed agent-submitted final answers with measured completion and tool use, the proportion completing less than 25\% of local goals while retaining more than 50\% of the tool allowance increases from 2.4\% at 64K to 15.3\% at 256K and 54.0\% at 1M, as shown in Figure~\ref{fig:sustained-progress}b. At the largest workload, more than half of these failed finals therefore leave most local goals unfinished without exhausting the tool-call budget. Appendix~\ref{app:joint-analysis} reports the counts and threshold sensitivity.

\ifdefined\SkillHorizonArxiv\FloatBarrier\else\clearpage\fi
\subsection{From Local Execution to Global Delivery}
\begin{figure}[!htb]
\centering
\includegraphics[width=.90\linewidth]{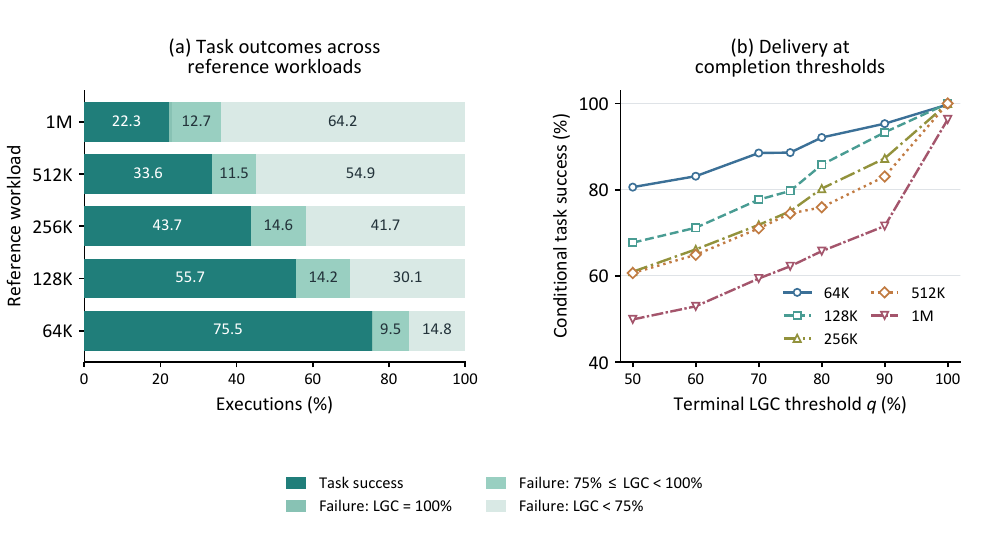}
\caption{Local completion and global delivery across seven environments and five models. (a) Joint outcomes for all 5,250 executions, with 1,050 per workload band. (b) Task success among executions whose terminal LGC meets threshold $q$, shown separately for each workload band. Table~\ref{tab:threshold-delivery} reports the numerators and denominators.}
\label{fig:local-delivery}
\end{figure}

\textbf{Substantial local progress does not guarantee complete task delivery.} Across seven environments, 656 executions complete at least three-quarters of their local goals but leave the remaining work unfinished. These executions account for 9.5--14.6\% of tasks within each workload band. Among executions ending with LGC of at least 80\%, task success falls from 92.1\% at 64K to 65.7\% at 1M. Even at a 90\% completion threshold, success falls from 95.3\% to 71.6\%. Thus, substantial local progress is less consistently accompanied by complete delivery at larger workloads.

\textbf{Successful local computation does not ensure valid downstream use.} In SocialOps, downstream submissions can reproduce the outputs of successful skill-script executions and still be rejected. The task-averaged rejection rate of these submissions rises from 3.4\% at 64K to 16.5\% at 1M. Most rejections concern historical-source requirements or current audience loads. These failures occur despite agreement between the submitted values and script outputs, showing that skill composition also depends on whether those outputs satisfy the receiving operation's business constraints.

\FloatBarrier
\subsection{Recovery over Long Horizons}
\begin{figure}[!htb]
\centering
\includegraphics[width=.92\linewidth]{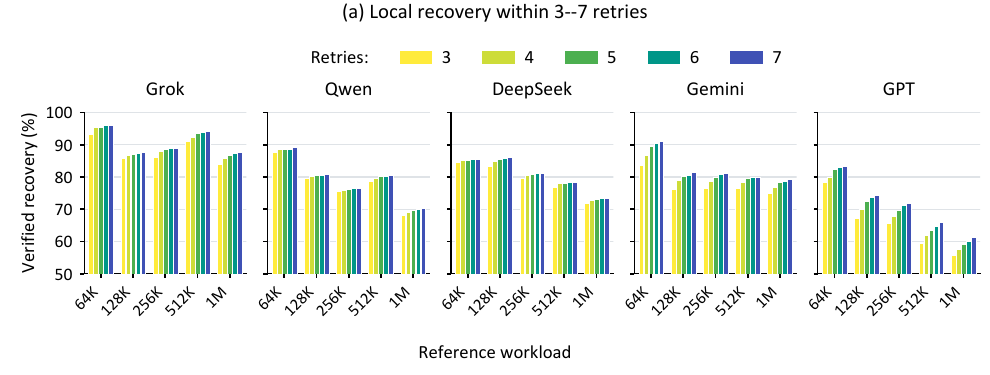}
\par\vspace{2pt}
\includegraphics[width=.92\linewidth]{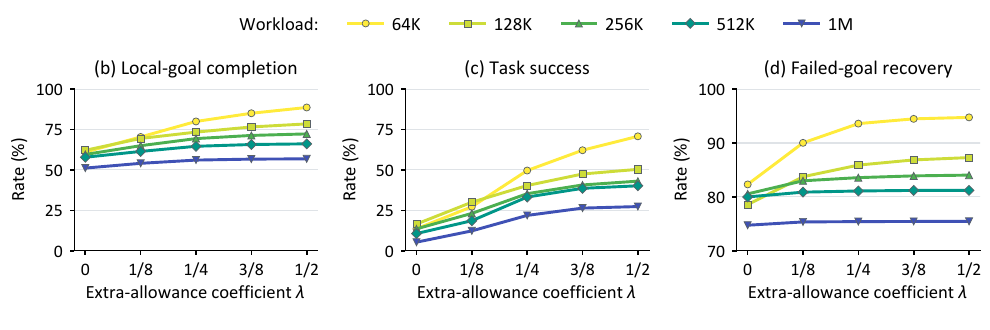}
\caption{Recovery across workload scales. (a) Task-averaged recovery within 3--7 subsequent checks of the same goal and obligation cycle, shown by model and workload. (b--d) Local-goal completion, task success, and failed-goal recovery along fixed trajectories at $B_\lambda=R+\lceil\lambda A\rceil$, using the same 3,374 executions with public failures by $R$ and retaining earlier terminal states. Appendix~\ref{app:goal-followup} details the populations and counting rules.}
\label{fig:recovery-combined}
\end{figure}

\textbf{Local recovery becomes less reliable as workloads grow.} Figure~\ref{fig:recovery-combined}a shows recovery within three to seven subsequent verification attempts for each model across the five workload bands. Extending this range from three to seven raises task-averaged recovery from 84.4\% to 87.9\% at 64K and from 69.6\% to 73.0\% at 1M, leaving a gap of approximately 15 percentage points. All five models exhibit lower recovery at 1M under both limits. Even with seven retries, the decline from 64K to 1M ranges from 8.4 percentage points for Grok to 22.0 points for GPT.

\textbf{Continued execution yields smaller completion gains at longer horizons.} Among executions with a business failure within the reference allowance, extending the observed budget prefix from $\lambda=0$ to $\lambda=1/2$ raises LGC from 61.0\% to 88.6\% at 64K, but from 51.2\% to 56.9\% at 1M. Task success increases from 13.6\% to 70.8\% at 64K and from 5.3\% to 27.4\% at 1M. Thus, longer prefixes capture additional complete deliveries at both scales, while the gain in local-goal completion is substantially smaller at 1M. Figure~\ref{fig:recovery-combined}b--d reports completion, delivery, and failed-goal recovery for these executions at $B_\lambda=R+\lceil\lambda A\rceil$, retaining earlier terminal states. Appendix~\ref{app:budget-prefix} details the analysis.

\ifdefined\SkillHorizonArxiv\else\FloatBarrier\fi
\subsection{Dependency Continuity across Compaction}
\begin{wraptable}[15]{r}{.52\textwidth}
\vspace{-8pt}
\centering\small
\setlength{\tabcolsep}{3pt}
\textbf{Dependency continuity}\par\smallskip
\begin{tabularx}{\linewidth}{@{}Y>{\centering\arraybackslash}p{1.24cm}>{\centering\arraybackslash}p{1.24cm}@{}}
\toprule
Metric & Retained & Removed \\
\midrule
First-check acceptance (\%) & 84.5 & 77.2 \\
Ever accepted (\%) & 91.8 & 89.8 \\
Additional checks & 0.17 & 0.34 \\
\bottomrule
\end{tabularx}
\par\vspace{4pt}
\textbf{Task-level outcomes}\par\smallskip
\begin{tabularx}{\linewidth}{@{}Y>{\centering\arraybackslash}p{1.24cm}>{\centering\arraybackslash}p{1.24cm}@{}}
\toprule
Metric & Without & With \\
\midrule
LGC (\%) & 37.0 & 71.3 \\
TS (\%) & 23.4 & 41.1 \\
Mean tool calls & 303 & 1,159 \\
\bottomrule
\end{tabularx}
\caption{Local dependency checks and task-level outcomes across compaction.}
\label{tab:compaction-summary}
\vspace{-8pt}
\end{wraptable}

\textbf{Cross-compaction dependencies are associated with lower initial reliability and greater recovery effort.} Across seven environments, first-check acceptance is 77.2\% when required historical outputs have been removed from the retained context, compared with 84.5\% when they remain verbatim available. As shown in Table~\ref{tab:compaction-summary}, subsequent attempts narrow the acceptance gap to 1.9 percentage points, but average additional checks increase from 0.169 to 0.344. Recovery therefore closes much of the initial gap while requiring greater verification effort.

\textbf{Executions with compaction show greater observed progress.} With equal weights over 95 shared environment--workload--model strata, the 1,734 executions with compaction achieve 71.3\% local-goal completion and 41.1\% task success, compared with 37.0\% and 23.4\% for 1,116 executions without compaction (Table~\ref{tab:compaction-summary}). They also make more tool calls, averaging 1,159 versus 303. Compaction exposure is endogenous: early stopping can prevent it, while continued execution creates opportunities for it. These associations do not establish a causal benefit of compaction. Appendix~\ref{app:required-dependency-compaction} reports environment-level results and sample sizes.

\ifdefined\SkillHorizonArxiv\FloatBarrier\else\clearpage\fi
\subsection{Task-Sampling Sensitivity}
\begin{wrapfigure}{r}{.48\textwidth}
\vspace{-20pt}
\centering
\setlength{\abovecaptionskip}{2pt}
\includegraphics[width=\linewidth]{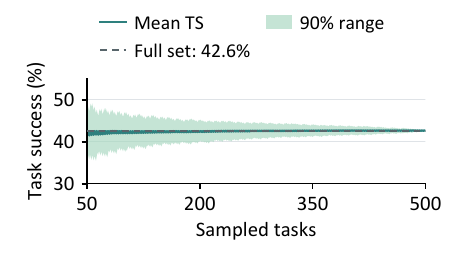}
\caption{Mean and central 90\% range across 1,000 family-balanced task orders.}
\label{fig:task-sampling}
\vspace{-8pt}
\end{wrapfigure}

\textbf{Success estimates remain close across family-balanced samples of the evaluated task pool.} We evaluate Gemini 3.8 Flash on 500 reference-validated FinanceOps tasks spanning nine configurations at 128K. Across 1,000 sampled orders, mean task success stays close to the full-set rate of 42.6\%, while the central 90\% sampling range narrows from 36.0--48.0\% at 50 tasks to 41.0--44.3\% at 300, as shown in Figure~\ref{fig:task-sampling}. Appendix~\ref{app:task-sampling} details the sampling procedure.

\par
\makeatletter
\ifnum\c@WF@wrappedlines>1
  \@tempcnta=\c@WF@wrappedlines
  \advance\@tempcnta by -1
  \dimen@=\baselineskip
  \multiply\dimen@ by \@tempcnta
  \vskip\dimen@
\fi
\WFclear
\makeatother
\FloatBarrier
\section{Related Work}
\label{sec:related}

\paragraph{Long-horizon benchmarks.}
AgentLongBench studies information synthesis over extended histories \citep{fang2026agentlongbench}, while LOCA-bench scales environment states to increase context without changing task semantics \citep{zeng2026loca}. WorkArena++ and Toolathlon evaluate compositional workflows and cross-application execution \citep{boisvert2024workarena,li2025toolathlon}. AgencyBench assesses extended real-world tasks through deliverables and rubrics \citep{li2026agencybench}, while Long-Horizon-Terminal-Bench provides subtask-level grading \citep{li2026longhorizonterminal}. \bench{} focuses on expanding skill-grounded business activities, artifact dependencies, and shared-state constraints to evaluate local progress and whole-task delivery as requirements grow.

\paragraph{Agent skills.}
Reusable skills encode procedures that agents can apply across tasks. Voyager accumulates and composes executable skills \citep{wang2023voyager}, while the Agent Skills specification packages instructions with optional scripts and reference resources \citep{agentskills_spec}. SkillsBench compares task performance without skills, with curated skills, and with self-generated skills \citep{li2026skillsbench}. AgentSkillOS combines hierarchical skill retrieval with DAG-based orchestration \citep{li2026agentskillos}. These works study skills for task solving. \bench{} instead uses skills to define business activities and task dependencies, leaving execution plans to the agent.

\paragraph{MCP-based environments and benchmarks.}
MCP standardizes access to external tools and data sources \citep{mcp2025spec}. MCP-Universe, MCP-Bench, and MCPMark evaluate agents through real-world tool integrations, multi-step coordination, and programmatic outcome checks \citep{luo2025mcpuniverse,wang2025mcpbench,wu2025mcpmark}. Related API-based benchmarks, including AppWorld and $\tau$-bench, assess task completion through application-state and database-goal checks \citep{trivedi2024appworld,yao2024taubench}. However, expanding these task suites still involves task-specific generation, review, or verifier design, leaving systematic variation of workload and dependencies less directly supported.

\begin{samepage}
\section{Conclusion}
We introduced \bench{}, a benchmark that constructs long-horizon business tasks by composing skill-grounded activities and their dependencies across seven executable environments. Configurable generators expand required work and its connections, while public reference executions establish feasibility and programmatic checks evaluate business outcomes. Across five models and 64K--1M reference workloads, effective progress falls behind growing task requirements. Trace analyses reveal gaps between successful local execution and accepted business delivery, together with heterogeneous recovery and cross-compaction dependency-use patterns. These findings highlight the importance of evaluating sustained progress and downstream validity alongside local execution success.

\bench{} provides a configurable testbed for studying how context management, skill use, and execution strategies support complete business workflows. Its generators support evaluation on new instances and larger workloads to assess whether improvements transfer across task scales.
\par
\label{main:end}
\end{samepage}

\clearpage
\endgroup
\section*{Ethics Statement}
Tasks use synthetic objects in isolated environments, with no changes submitted to real business services. Private evaluation data are separated from public resources. Raw execution records preserve operational events independently of the paper's outcome aggregation. External code and resources retain their provenance and applicable licensing requirements.

\bibliography{references}
\bibliographystyle{iclr2027_conference}
\clearpage
\appendix
\ifdefined\SkillHorizonArxiv
  \clubpenalty=10000
  \widowpenalty=10000
  \RecustomVerbatimEnvironment{PaperPrompt}{Verbatim}{fontsize=\footnotesize,
    breaklines=true,breakanywhere=true,samepage=false,frame=single,framesep=5pt}
\fi
\section{Environment and Task Construction}
\label{app:length}

\subsection{Environment Resources and Instantiation}
\label{app:environment-resources}
FinanceOps and FieldOps generate business objects from parameterized domain
models. The other five environments adapt seeded backend assets from
Infinite Context Compact \citep{infinitecontext}, described as synthetic or
mixed, and add generated task-specific objects, dependencies, and updates.
Table~\ref{tab:environment-resources} summarizes these resources and the
public skill libraries.

\begin{table}[!htbp]
\centering\small
\caption{Environment resources and public skill libraries. Seed records include relationship and event records in the underlying databases. Generated resources are instantiated from task parameters. Table~\ref{tab:workload-growth} reports the business scale of evaluated tasks.}
\label{tab:environment-resources}
\setlength{\tabcolsep}{4pt}
\begin{tabularx}{\linewidth}{@{}lYrr@{}}
\toprule
Environment & Representative resources & Seed records & Skills \\
\midrule
FinanceOps & Financial sources, transactions, identity maps, invoices, and remittances & Generated & 12 \\
FieldOps & Devices, service networks, maintenance windows, crews, and spare parts & Generated & 9 \\
SocialOps & 4,200 posts; 180 campaigns; 900 assets & 20,734 & 8 \\
LogisticsOps & 850 passengers; 320 flights; 4,100 bookings & 21,150 & 9 \\
ResearchOps & 1,800 papers; 900 authors; 2,000 citations & 11,160 & 7 \\
DocOps & 140 projects; 1,600 pages; 6,400 document blocks & 22,612 & 6 \\
FactoryOps & 18 robot cells; 180 programs; 1,800 waypoints & 13,438 & 7 \\
\bottomrule
\end{tabularx}
\end{table}

\paragraph{Procedural resources.}
\label{app:skill}
Each skill package contains a \skillfile{} document specifying applicability,
required inputs, a procedure, and expected effects, with optional scripts
and references \citep{agentskills_spec}. Supplied scripts compute over
inputs selected and provided by the agent. Domain rules instantiate these
procedures over concrete objects and connect their outputs to receiving
operations. Appendix~\ref{app:construction-example} illustrates how a skill's
input requirements connect to downstream business goals.

\subsection{Domain-Specific Stage Construction}
\label{app:stage-construction}
\label{tab:stage-construction}
The following rules connect required business outcomes through upstream
artifacts and shared constraints; agents choose legal operation sequences.

\paragraph{Financial sources and stage closures.}
FinanceOps adds source work packages within logical stages. Its families require format and identity selection, evidence-backed component adjustments, or remittance reconciliation against original invoices and credit evidence. Stage closures combine all required source views with the preceding closure; review and publication cover the whole release. Checks assess numerical results and identity or payment bindings. Source corrections invalidate affected outputs and downstream closures while preserving immutable historical views and unrelated valid work. Business facts vary independently of opaque identities; public decision metadata supports input selection.

\paragraph{Maintenance and service commissioning.}
FieldOps expands device-maintenance and service-commissioning networks. Upstream commissioning records supply pressures, flows, identities, and revisions that constrain downstream inlet conditions and maintenance windows. Repairs and commissioning require qualified crews, compatible spares, and shared flow capacity. Configurations vary devices per service, parent joins, and physical requirements. The selected set has 75 normal and 75 recovery tasks. Recovery scenarios introduce two progress-triggered supply changes or a local device fault, requiring renewed certification or revised repairs and downstream certifications. Valid repairs and unrelated work remain reusable. Corrective re-isolation checks scheduling and dependencies before reopening restored boundaries.

\paragraph{Campaign waves.}
SocialOps links campaign waves through measured reports. Staggered, distributed, alternating fork--join, and protected-branch patterns vary parent origins and joins. New variants combine parent winners' audiences and calls to action with arm-specific permissions and newly consented audiences. Each wave adds publication, measurement, reporting, and deployment obligations under scheduling and exposure conditions. Rights updates require affected unpublished drafts to obtain valid assets and approvals while preserving published history.

\paragraph{Itinerary and custody chains.}
LogisticsOps adds route-leg bookings and baggage deliveries. Cohorts share ground transfers; staggered or distributed patterns vary delivery spans and joins. Each freight movement consumes a release or predecessor manifest establishing custody at its departure hub and produces the next handoff. Route-specific services, schedules, handling permissions, and weight-based shared capacities constrain feasible connections. Completion requires both passenger arrangements and baggage delivery. Carrier withdrawals require affected custody plans to be revised while valid arrangements are retained.

\paragraph{Calibration and evidence rounds.}
ResearchOps adds calibration anchors and studies by acquisition round. Derived anchors compose earlier calibration maps, and at least one compatible study uses each new anchor. Contributions bind current annotations for the named calibration and its ancestors. Metric, population, and unit rules determine usable evidence. Corrections revise derived interpretations and withdrawals change eligible contributions; prior committed reviews retain their historical identity.

\paragraph{Document stages.}
DocOps links each page to its previous-stage counterpart. Chain--join adds periodic cross-lane parents, braided adds eligible cross-lane parents at each level, and distributed also adds older parents. Local grants specify purpose and expiry; parent contracts enumerate consumers, weights, and allocation limits. Publication combines parent contracts with local fields under schema, purpose, expiry, and capacity rules. Public bulletins announce amendments, preserving an unaffected lane and valid history. Page acceptance requires a current publication and matching tracker binding. In reference executions, 200 updates across 100 tasks each affect 2--29 published pages while preserving 3--56 others.

\paragraph{Production levels.}
FactoryOps assigns each order a preceding-level supplier and another compatible supplier from the preceding, an earlier, or the root level. Every nonterminal order supplies later work. Terminal demand propagates backward through the bill of materials to determine upstream quantities and raw stock. Produced units carry lineage and quality measurements downstream; completion requires terminal dispatch. Equipment occupancy, maintenance, calibration, and selective rework constrain execution. Tooling drift affects future work while accepted units retain their status.

\subsection{Generation and Workload Calibration}
\label{app:generation-rules}
Each candidate specifies a workflow family, business-scale parameters,
dependency pattern where applicable, and seed. Its domain builder generates
the objects, dependencies, initial resources, updates, and delivery
requirements. Successful public reference executions establish feasibility
and supply the workload measurement for band assignment. Evaluation seeds
are disjoint from development seeds.

\paragraph{Calibration and selection.}
Calibration adjusts business size under the fixed factors in
Table~\ref{tab:construction-recipes}, measures a successful public reference
execution, and assigns the resulting task to its workload band. Selection
uses reference feasibility and workload, independently of evaluated-model
outcomes. Per-task configurations in the released data specify the selected
parameters and seeds.

\begin{table}[htbp]
\centering\small
\caption{Environment-specific workload calibration. Fixed factors apply within each calibration search. Appendix~\ref{app:stage-construction} specifies dependency and update semantics.}
\label{tab:construction-recipes}
\begin{tabularx}{\linewidth}{@{}p{1.75cm}YY@{}}
\toprule
Domain & Fixed construction factors & Scale adjustment \\
\midrule
FinanceOps & Workflow family, variant, stage count, and seed & Adjust source width using public reference workload. \\
\addlinespace
FieldOps & Profile and scenario & Calibrate service and device counts with the public reference. \\
\addlinespace
SocialOps & Scenario and dependency pattern & Adjust campaigns, waves, and campaigns in the final wave using the fixed public reference policy. \\
\addlinespace
LogisticsOps & Family, topology, and route legs & Adjust cohort count; use the next planned seed only when the size granularity cannot fit the target interval. \\
\addlinespace
ResearchOps & Family, seed, sections, rounds, aliases, exclusions, and topology & Adjust studies per section and trim terminal studies while retaining at least one original study per topic. \\
\addlinespace
DocOps & Family, seed, and topology & Adjust document count $n$; derive depth as $\lceil n/3\rceil$. \\
\addlinespace
FactoryOps & Family, seed, levels, topology, and terminal quantities & Adjust orders per level; recompute upstream quantities from the generated supply graph. \\
\bottomrule
\end{tabularx}
\end{table}

\subsubsection{Length Bands and Counting}
\begin{table}[htbp]
\centering\small
\caption{Reference-workload bands, with an exclusive lower bound and inclusive upper bound. K denotes $1024$ tokens and M denotes $1024^2$ tokens. Each instance retains its measured reference length.\looseness=-1}
\begin{tabular}{@{}lrr@{}}
\toprule
Band & Lower bound & Upper bound \\
\midrule
64K & 0 & 65,536 \\
128K & 65,536 & 131,072 \\
256K & 131,072 & 262,144 \\
512K & 262,144 & 524,288 \\
1M & 524,288 & 1,048,576 \\
\bottomrule
\end{tabular}
\end{table}

We measure reference workload by counting tokens in public tool calls and returned observations, including tool names and arguments. Calls and results are serialized separately as compact UTF-8 JSON with sorted keys and tokenized using tiktoken's \texttt{o200k\_base} encoding \citep{openai_tiktoken}. Each occurrence is counted once, including repeated queries. The count excludes initial instructions, tool schemas, closing narration, compaction summaries, retransmitted request prefixes, hidden reasoning, and fixed-script internals.

\paragraph{Query strategy.}
Reference policies operate on publicly observed state, reuse valid observations, and query additional information as needed. After state changes, they reassess affected dependencies before reusing prior results.

Band assignment is fixed by the reference execution. An evaluated agent
can take a different path, and compaction can reduce its active context
while the task remains in the same reference-workload band.

\subsection{Business Scale across Workload Bands}
\label{app:construction-profiles}
Table~\ref{tab:workload-growth} shows how work counts, workflow dimensions,
and reference business calls change across the five workload bands.
Work counts describe the objects to process; stages, waves, rounds, and
levels describe their organization under the domain rules above.

\begin{table}[!htbp]
\centering\small
\caption{Business scale across workload bands. Entries are minimum--maximum ranges over 30 tasks per environment and band. Business calls count reference business-execution invocations, including each supplied script invocation as one call.}
\label{tab:workload-growth}
\setlength{\tabcolsep}{3pt}
\begin{tabular*}{\linewidth}{@{\extracolsep{\fill}}llrrrrr@{}}
\toprule
Environment & Attribute & 64K & 128K & 256K & 512K & 1M \\
\midrule
FinanceOps & Work packages & 24--64 & 54--132 & 110--270 & 224--544 & 480--1,152 \\
 & Logical stages & 4 & 6 & 10 & 16 & 24 \\
 & Business calls & 126--300 & 249--622 & 509--1,228 & 1,033--2,438 & 2,221--5,092 \\
\midrule
FieldOps & Devices & 9--33 & 30--64 & 49--128 & 120--294 & 294--768 \\
 & Services & 8--20 & 16--36 & 32--68 & 64--196 & 176--446 \\
 & Business calls & 139--206 & 301--411 & 573--728 & 1,281--1,765 & 3,593--4,040 \\
\midrule
SocialOps & Posts & 9 & 15 & 27--30 & 48--57 & 90--111 \\
 & Waves & 2 & 3 & 3--4 & 6--7 & 10--13 \\
 & Business calls & 109--117 & 180--188 & 316--361 & 563--672 & 1,055--1,306 \\
\midrule
LogisticsOps & Bookings & 16 & 32 & 60--70 & 120--144 & 224--320 \\
 & Legs & 4 & 4 & 5 & 6 & 8 \\
 & Business calls & 122--139 & 248--288 & 519--623 & 1,074--1,187 & 2,232--2,560 \\
\midrule
ResearchOps & Source works & 15--21 & 39 & 76--92 & 165--185 & 365--395 \\
 & Rounds & 3 & 3 & 4 & 4--5 & 5--6 \\
 & Business calls & 183--232 & 379--434 & 784--900 & 1,635--1,825 & 3,498--3,850 \\
\midrule
DocOps & Pages & 6--10 & 10--19 & 19--36 & 31--63 & 52--105 \\
 & Configured depth & 2--4 & 4--7 & 7--12 & 11--21 & 18--35 \\
 & Business calls & 73--99 & 121--195 & 225--355 & 385--603 & 681--1,083 \\
\midrule
FactoryOps & Orders & 12--14 & 20--28 & 48--60 & 65--120 & 140--272 \\
 & Levels & 2 & 3--4 & 3--5 & 4--5 & 4--5 \\
 & Business calls & 109--132 & 173--276 & 413--598 & 586--1,163 & 1,214--2,373 \\
\bottomrule
\end{tabular*}
\end{table}

\subsection{Skill Inputs and Downstream Dependencies}
\label{app:construction-example}
FinanceOps illustrates how generated business requirements connect reusable
skills. Each source specifies a format contract, an authoritative identity
mapping, and a rule for conflicting revisions. The agent selects these
inputs for \texttt{normalize-source}; \texttt{aggregate-source} consumes the
normalized records to produce entity totals. A stage closes only when its
source outputs and the preceding stage's closure are current. The release
then requires every stage's closure, review, and publication. Thus a local
input-selection error can propagate through aggregation to release-level
validation.

\label{app:skill-package-example}
The normalization package contains a \skillfile{} document and an arithmetic
script. Its instructions specify how the agent should select inputs:
\begin{PaperPrompt}
source_id, contract_id, identity_source_id, identity_map_id, conflict_policy. Read the authoritative identity_source_id declared by the source, even if it was released much earlier. Select the contract matching the source format and its effective period. Conflicting order revisions follow conflict_policy: first preserves the first record; latest_revision applies sale/correction rows, ignores metadata-only rows, and excludes withdrawn orders. A metadata row never changes an earlier financial amount. A superseding identity mapping applies only in its effective range. Normalized output cites mapping-selection rules and only the selected mapping/account values it uses; changing a different mapping does not invalidate it. Wrong but well-typed choices may create incorrect proposals. Output contains canonical entity and integer minor-unit amounts.
\end{PaperPrompt}
The script computes over the supplied arguments. A well-typed choice can
produce a proposal even when it uses an incorrect contract or mapping.
Terminal evaluation checks its business correctness alongside the input
and version checks enforced during execution.

\FloatBarrier
\section{Agent Protocol and Prompts}
\label{app:protocols}

\subsection{Execution Settings}
All models receive the same tasks. Requests return one non-streaming response, with temperature and top-$p$ unspecified; Table~\ref{tab:request-settings} gives generation settings.
\begin{table}[htbp]
\centering\small
\caption{Requested per-stage generation settings. Output limits include reported reasoning tokens where applicable.}
\label{tab:request-settings}
\begin{tabularx}{\linewidth}{@{}lYr@{}}
\toprule
Model identifier & Reasoning setting & Max. output tokens \\
\midrule
\path{grok-4.6} & Medium & 8,192 \\
\path{qwen3.8-flash} & Medium or 4,096 tokens & 8,192 \\
\path{deepseek-flash} & Medium & 32,768 \\
\path{gemini-3.8-flash} & Medium & 8,192 \\
\path{gpt-5.6-sol} & Medium & 8,192 \\
\bottomrule
\end{tabularx}
\end{table}

Within an action batch, calls execute sequentially until the first public failure or batch completion; unattempted calls are not charged. Queries, skill reads, business operations, script invocations, and identifiable rejected actions each consume one tool attempt, subject to declaration exemptions. Summary requests and transport retries do not consume tool attempts.

The allowance is shared across skills and continuations. Exhaustion stops execution after the last permitted attempt is evaluated; voluntary termination follows the confirmation prompt below.

\subsection{Prompts}
\label{app:prompts}
\paragraph{Actor and access instructions.}
The actor instruction below is from an evaluated DocOps request.
\begin{PaperPrompt}
Complete the user's business project through the public tools below.
Return exactly one JSON object, without markdown. To call tools return
{"type":"actions","actions":[{"name":"TOOL","arguments":{},"skill_run_id":"OPTIONAL_LABEL"}]}.
You may issue 1 to 32 independent calls per response. All arguments must be explicit
and fixed before receiving results. To use a returned value, wait for the next turn.
No code, expressions, references to future results, or controller-generated arguments
are executed.
Use {"type":"final","text":"your final answer"} to request termination
of the entire task, whether complete or incomplete.
Accurately state what was completed and what remains unfinished.
If incomplete, explain why you are ending based on the public
information available to you.
Use actions to continue working; final is not a progress update.
Read skill instructions using read_skill_file. Fixed scripts are available only via
run_skill_script. There is no terminal, arbitrary code execution or personal workspace.
begin_skill declares a reusable skill; finish_skill completed is your claim of local
completion, not business certification. skill_run_id is optional envelope metadata,
not a business argument. Business state and artifacts persist through public tools.
No hidden evaluator feedback is available.
\end{PaperPrompt}
Public schemas and the tool budget accompany the system instruction; tasks and tool results use user-role messages. The parser accepts fenced action JSON with up to 1,024 characters of surrounding commentary on each side; final responses require bare JSON.

\paragraph{Configurable task entry.}
All seven environments support two task-entry modes. Both expose the same public task brief and skill catalog through \texttt{read\_task\_brief}; current business state is accessed through domain tools. The examples below present the same four-page DocOps task in both modes. Brief excerpts show the goal, requested pages, and skill names; both modes share the actor and skill-use instructions.
\par\medskip\noindent
\Needspace{8\baselineskip}
\textbf{Initial brief: task supplied before the first response.}
\par\nobreak
\begin{PaperPrompt}
[Entry instruction]
Complete the DocOps business task using public tools.
The task brief is supplied below and can be reread with read_task_brief.

[Initial user message: public task brief excerpt]
Goal:
Reconcile, review and publish the interface contracts for every requested PRD/RFC, including downstream joins. Preserve declared source fields and limits, use current parent publications, obtain actual engineering and security signoffs, publish accepted decisions and synchronize each tracker to its own current publication. Requirements are in page blocks. Legal alternative ordering, prose and recovery are allowed. After the initial delivery, poll the planning change window and reconcile any announced amendments before final delivery. Respect shared consumer capacity reservations across teams and nonadjacent stages. Publish actual quota values, not just dependency identities. Preserve each original quota owner's purpose restrictions and expiry through all downstream reservations.

Requested pages (page / project / tracker):
page-001 / project-001 / database-item-001
page-002 / project-002 / database-item-002
page-003 / project-001 / database-item-003
page-004 / project-002 / database-item-004

Skills:
coordinate-document-review
publish-reviewed-decision
reconcile-interface-contracts
revise-planning-document
synchronize-planning-trackers
trace-planning-dependencies
\end{PaperPrompt}

\par\medskip\noindent
\Needspace{8\baselineskip}
\textbf{Tool discovery: task obtained through a public call.}
\par\nobreak
\begin{PaperPrompt}
[Entry instruction]
Complete the DocOps business task using public tools.
Obtain the task brief by calling read_task_brief with empty arguments.

[Assistant action]
{"type":"actions","actions":[{"name":"read_task_brief","arguments":{}}]}

[Public tool result: task brief excerpt]
Goal:
Reconcile, review and publish the interface contracts for every requested PRD/RFC, including downstream joins. Preserve declared source fields and limits, use current parent publications, obtain actual engineering and security signoffs, publish accepted decisions and synchronize each tracker to its own current publication. Requirements are in page blocks. Legal alternative ordering, prose and recovery are allowed. After the initial delivery, poll the planning change window and reconcile any announced amendments before final delivery. Respect shared consumer capacity reservations across teams and nonadjacent stages. Publish actual quota values, not just dependency identities. Preserve each original quota owner's purpose restrictions and expiry through all downstream reservations.

Requested pages (page / project / tracker):
page-001 / project-001 / database-item-001
page-002 / project-002 / database-item-002
page-003 / project-001 / database-item-003
page-004 / project-002 / database-item-004

Skills:
coordinate-document-review
publish-reviewed-decision
reconcile-interface-contracts
revise-planning-document
synchronize-planning-trackers
trace-planning-dependencies
\end{PaperPrompt}
\par\medskip

Explicit reads consume tool attempts; initial display consumes none. In both modes, the brief enters ordinary history subject to compaction.

\paragraph{Final-answer check and confirmation.}
\begin{PaperPrompt}
Before choosing final, compare the user's goal with the public results you have
actually observed. State remaining work or an observed blocker accurately.
\end{PaperPrompt}
A final submission triggers:
\begin{PaperPrompt}
Are you sure you want to end the entire task and submit the current result?

Confirming will submit the current state for evaluation, after which no further actions can be taken. To confirm, submit final again. To continue working, submit actions.

After any actual non-declaration tool attempt, including a rejected attempt, a later final will require a new confirmation. Existing budgets and time limits still apply.
\end{PaperPrompt}
\paragraph{Summary instruction and history replacement.}
\begin{PaperPrompt}
Summarize the model-visible business conversation for continuation.
Preserve the user's goal, skill decisions, actual artifact IDs and dependencies,
source revisions, notices, errors, unresolved obligations and next intended steps.
Distinguish verified tool results from your guesses. Do not invent task progress.
Return plain text only, do not call tools. This summary replaces older conversation;
the public initial instructions and any separately retained recent turns remain available.
\end{PaperPrompt}
Compaction starts at 200,000 estimated input tokens, with a local capacity guard of 480,000 \citep{anthropic2025context}. The evaluated model uses its per-stage response limit. Initial instructions and four, two, or zero recent messages are retained as complete action--result pairs, keeping instructions plus retained dialogue below 64,000 tokens. Older dialogue is replaced by a user-role summary; business state persists. Estimates use \texttt{o200k\_base} plus $256+16m$ tokens for $m$ messages. Adoption requires a complete, nonempty summary and a smaller input: full-prefix replacement must fall below 200,000 tokens, while staged replacement retains unprocessed history for later passes.

\FloatBarrier
\section{Evaluation Definitions}
\label{app:metrics}
\label{app:obligations}

\subsection{Terminal Business Outcomes}
For $N$ model--task executions, let $s_i\in\{0,1\}$ indicate whether execution $i$ satisfies all final business requirements. Let $g_i$ be the number of local goals satisfied at termination and $G_i$ the number required by the full task, including goals in unattempted stages or rounds. We compute
\begin{equation}
\mathrm{TS}=\frac{100}{N}\sum_{i=1}^{N}s_i,
\qquad
\mathrm{LGC}=\frac{100}{N}\sum_{i=1}^{N}\frac{g_i}{G_i}.
\label{eq:outcome-metrics}
\end{equation}
Both metrics cover all 5,250 model--task executions, with equal weight per execution. Table~\ref{tab:native-count-coverage} defines the environment-specific goals and delivery requirements. These scoring units can differ from construction units: SocialOps counts completed briefs rather than posts, and ResearchOps counts accepted rounds rather than source works.

\begin{table}[!htbp]
\centering\small
\caption{Local-goal units, completion checks, and whole-task delivery requirements. TS requires every local goal and the additional requirements in the final column.}
\label{tab:native-count-coverage}
\begin{tabularx}{\linewidth}{@{}l>{\raggedright\arraybackslash}p{1.7cm}YY@{}}
\toprule
Environment & Local-goal unit & Local completion & Additional task requirements \\
\midrule
FinanceOps & Source work package & Correct current outputs and required historical views incorporated into stage closures & Stage accounting, release review, publication, and released obligations \\
FieldOps & Device case & Restored device with passing current tests & Service commissioning, closed isolations, resource accounting, and system acceptance \\
SocialOps & Campaign brief & Accepted experiment, report, and deployment with valid content, audience, and source evidence & Scheduled updates, shared publication constraints, and business-input integrity \\
LogisticsOps & Booking & Acceptance by network evaluation & Cargo and freight delivery requirements \\
ResearchOps & Review round & Accepted review with valid evidence and synthesis & Completion of all rounds and event/state integrity \\
DocOps & Page & Accepted current page, publication, and matching tracker evidence & Completed change windows and update/state integrity \\
FactoryOps & Production order & Required quantity of quality-accepted output & Terminal dispatch and material, equipment, and execution integrity \\
\bottomrule
\end{tabularx}
\end{table}

\paragraph{Local and global completion.}
\label{app:claims}
Each required object--goal pair is evaluated once under its applicable version. The denominator includes all required local goals and applicable conditional goals. ResearchOps uses the stored creation-state evaluations of committed reviews for earlier rounds and reevaluates the current round against its terminal state. FactoryOps counts quantity-complete intermediate orders as well as terminal orders. Task success requires all local goals and the remaining delivery, integrity, and update requirements.

\subsection{Execution and Recovery Events}
\label{app:artifact-flow}
Dependencies are traced through the identities, versions, and contents of artifacts produced and consumed by business operations. A successful script return records computation on the submitted inputs; consumer acceptance records the receiving operation's business checks. Task outcomes are scored from the final business state.

\label{sec:recovery}
A business-goal cycle groups attempts to satisfy the same checkable requirement for one object. Changes to parameters or replacement artifacts remain in that cycle until acceptance or the end of observation. Invalidation of an accepted goal starts a new update cycle. Failure correction begins with a failed business attempt; update recovery begins with invalidation. Both require a corrective action followed by verified local acceptance.

Recovery rates include unresolved eligible cycles. Public-feedback analyses start from observed business failures, excluding provider retries and retrospective content findings. Verification effort counts subsequent checks of the affected goal; intervening work on other goals is not counted as another check. Appendix~\ref{app:goal-followup} specifies the retry and budget-prefix comparisons, and Appendix~\ref{app:required-dependency-compaction} defines consumption-time input visibility.

\subsection{Aggregation and Measurement Populations}
\label{app:statistics}
Each model--task pair contributes one terminal outcome. Continuations form one execution, with shared prefixes counted once. Process rates are computed within eligible executions before aggregation; each analysis specifies its eligibility rules and aggregation weights. Absolute local-goal counts are reported separately by environment.

Stopping analyses use observed failed final submissions with the required progress and budget measurements. Dependency-visibility analyses reconstruct the actual consuming requests after adopted history replacements. Each analysis reports its selection criteria and sample sizes alongside the results.

\FloatBarrier
\section{Detailed Results and Costs}
\label{app:detailed-results}

\label{app:result-status}
Table~\ref{tab:detailed-results} reports LGC and TS by workload; Table~\ref{tab:dependency-families} reports both metrics for the twenty task families, pooling the five workload bands. These families vary jointly in mechanisms and scale. The 5,250 executions comprise 2,424 business successes (S), 2,657 unsuccessful non-budget endpoints (F), and 169 unsuccessful budget-exhausted endpoints (B). Voluntary-stopping analyses use actual final submissions.
\begin{table}[!tp]
\centering\small
\caption{Local-goal completion / task success (LGC / TS, \%) by environment and workload. Both metrics use all 30 executions per cell. LGC averages native terminal-goal fractions; TS requires \mbox{full business acceptance.}}
\label{tab:detailed-results}
\setlength{\tabcolsep}{3pt}
\begin{tabular*}{\linewidth}{@{\extracolsep{\fill}}llrrrrr@{}}
\toprule
Environment & Model & 64K & 128K & 256K & 512K & 1M \\
\midrule
\textbf{SocialOps} & Grok 4.6 & \scorepair{98.9}{96.7} & \scorepair{98.0}{93.3} & \scorepair{96.7}{96.7} & \scorepair{96.8}{93.3} & \scorepair{99.3}{93.3} \\
 & Qwen3.8-Flash & \scorepair{100.0}{100.0} & \scorepair{100.0}{100.0} & \scorepair{97.8}{96.7} & \scorepair{95.6}{90.0} & \scorepair{80.7}{53.3} \\
 & DeepSeek V4.1 Flash & \scorepair{100.0}{100.0} & \scorepair{98.7}{96.7} & \scorepair{97.8}{96.7} & \scorepair{96.9}{86.7} & \scorepair{90.3}{80.0} \\
 & Gemini 3.8 Flash & \scorepair{96.7}{90.0} & \scorepair{92.7}{63.3} & \scorepair{90.7}{76.7} & \scorepair{80.4}{50.0} & \scorepair{47.4}{30.0} \\
 & GPT-5.6 Sol & \scorepair{70.0}{53.3} & \scorepair{54.0}{26.7} & \scorepair{46.9}{13.3} & \scorepair{33.5}{10.0} & \scorepair{20.6}{0.0} \\
\midrule
\textbf{LogisticsOps} & Grok 4.6 & \scorepair{100.0}{100.0} & \scorepair{99.1}{96.7} & \scorepair{100.0}{100.0} & \scorepair{99.7}{96.7} & \scorepair{89.2}{83.3} \\
 & Qwen3.8-Flash & \scorepair{78.5}{70.0} & \scorepair{58.1}{30.0} & \scorepair{26.5}{10.0} & \scorepair{12.0}{3.3} & \scorepair{3.6}{0.0} \\
 & DeepSeek V4.1 Flash & \scorepair{77.5}{76.7} & \scorepair{70.5}{43.3} & \scorepair{51.4}{16.7} & \scorepair{30.1}{0.0} & \scorepair{12.8}{0.0} \\
 & Gemini 3.8 Flash & \scorepair{96.7}{93.3} & \scorepair{88.5}{80.0} & \scorepair{57.1}{43.3} & \scorepair{2.0}{0.0} & \scorepair{0.4}{0.0} \\
 & GPT-5.6 Sol & \scorepair{75.0}{23.3} & \scorepair{60.6}{10.0} & \scorepair{54.6}{6.7} & \scorepair{38.1}{6.7} & \scorepair{14.3}{0.0} \\
\midrule
\textbf{ResearchOps} & Grok 4.6 & \scorepair{98.9}{96.7} & \scorepair{97.8}{93.3} & \scorepair{90.8}{80.0} & \scorepair{41.3}{13.3} & \scorepair{22.3}{0.0} \\
 & Qwen3.8-Flash & \scorepair{95.6}{93.3} & \scorepair{84.4}{60.0} & \scorepair{70.0}{36.7} & \scorepair{34.7}{3.3} & \scorepair{18.7}{0.0} \\
 & DeepSeek V4.1 Flash & \scorepair{58.9}{30.0} & \scorepair{38.9}{6.7} & \scorepair{29.2}{6.7} & \scorepair{14.8}{0.0} & \scorepair{8.1}{0.0} \\
 & Gemini 3.8 Flash & \scorepair{86.7}{66.7} & \scorepair{62.2}{36.7} & \scorepair{43.3}{13.3} & \scorepair{17.8}{0.0} & \scorepair{6.2}{0.0} \\
 & GPT-5.6 Sol & \scorepair{76.7}{46.7} & \scorepair{37.8}{10.0} & \scorepair{18.3}{0.0} & \scorepair{10.2}{0.0} & \scorepair{3.7}{0.0} \\
\midrule
\textbf{DocOps} & Grok 4.6 & \scorepair{100.0}{100.0} & \scorepair{100.0}{100.0} & \scorepair{100.0}{100.0} & \scorepair{100.0}{100.0} & \scorepair{100.0}{100.0} \\
 & Qwen3.8-Flash & \scorepair{100.0}{100.0} & \scorepair{100.0}{100.0} & \scorepair{99.7}{90.0} & \scorepair{93.8}{76.7} & \scorepair{87.2}{53.3} \\
 & DeepSeek V4.1 Flash & \scorepair{100.0}{100.0} & \scorepair{99.5}{96.7} & \scorepair{98.7}{80.0} & \scorepair{99.6}{86.7} & \scorepair{97.2}{66.7} \\
 & Gemini 3.8 Flash & \scorepair{98.9}{96.7} & \scorepair{99.3}{96.7} & \scorepair{92.9}{86.7} & \scorepair{86.7}{83.3} & \scorepair{32.8}{13.3} \\
 & GPT-5.6 Sol & \scorepair{83.9}{66.7} & \scorepair{61.8}{16.7} & \scorepair{62.7}{3.3} & \scorepair{54.4}{0.0} & \scorepair{40.6}{0.0} \\
\midrule
\textbf{FactoryOps} & Grok 4.6 & \scorepair{98.1}{90.0} & \scorepair{93.8}{73.3} & \scorepair{81.8}{46.7} & \scorepair{60.5}{16.7} & \scorepair{56.3}{0.0} \\
 & Qwen3.8-Flash & \scorepair{90.0}{66.7} & \scorepair{58.3}{13.3} & \scorepair{44.9}{0.0} & \scorepair{31.7}{0.0} & \scorepair{9.2}{0.0} \\
 & DeepSeek V4.1 Flash & \scorepair{89.7}{56.7} & \scorepair{64.1}{10.0} & \scorepair{48.7}{0.0} & \scorepair{22.4}{0.0} & \scorepair{14.4}{0.0} \\
 & Gemini 3.8 Flash & \scorepair{78.7}{46.7} & \scorepair{52.0}{20.0} & \scorepair{26.2}{3.3} & \scorepair{5.5}{0.0} & \scorepair{11.5}{0.0} \\
 & GPT-5.6 Sol & \scorepair{71.1}{16.7} & \scorepair{44.0}{0.0} & \scorepair{33.9}{0.0} & \scorepair{16.9}{0.0} & \scorepair{7.8}{0.0} \\
\midrule
\textbf{FinanceOps} & Grok 4.6 & \scorepair{81.0}{56.7} & \scorepair{70.1}{36.7} & \scorepair{65.1}{36.7} & \scorepair{58.2}{30.0} & \scorepair{55.8}{20.0} \\
 & Qwen3.8-Flash & \scorepair{96.7}{53.3} & \scorepair{86.4}{43.3} & \scorepair{85.9}{33.3} & \scorepair{79.9}{33.3} & \scorepair{74.5}{20.0} \\
 & DeepSeek V4.1 Flash & \scorepair{96.0}{60.0} & \scorepair{88.5}{43.3} & \scorepair{89.3}{36.7} & \scorepair{89.8}{16.7} & \scorepair{85.0}{6.7} \\
 & Gemini 3.8 Flash & \scorepair{90.3}{60.0} & \scorepair{85.7}{46.7} & \scorepair{76.6}{36.7} & \scorepair{66.7}{36.7} & \scorepair{46.3}{13.3} \\
 & GPT-5.6 Sol & \scorepair{93.7}{66.7} & \scorepair{91.0}{16.7} & \scorepair{90.4}{0.0} & \scorepair{90.7}{3.3} & \scorepair{85.8}{0.0} \\
\midrule
\textbf{FieldOps} & Grok 4.6 & \scorepair{91.6}{90.0} & \scorepair{92.7}{76.7} & \scorepair{89.9}{63.3} & \scorepair{94.0}{73.3} & \scorepair{91.1}{70.0} \\
 & Qwen3.8-Flash & \scorepair{99.6}{96.7} & \scorepair{94.2}{83.3} & \scorepair{87.1}{50.0} & \scorepair{83.2}{70.0} & \scorepair{68.7}{33.3} \\
 & DeepSeek V4.1 Flash & \scorepair{100.0}{100.0} & \scorepair{93.4}{80.0} & \scorepair{92.3}{70.0} & \scorepair{93.4}{70.0} & \scorepair{84.7}{30.0} \\
 & Gemini 3.8 Flash & \scorepair{98.6}{96.7} & \scorepair{88.4}{66.7} & \scorepair{61.9}{50.0} & \scorepair{10.6}{6.7} & \scorepair{4.6}{0.0} \\
 & GPT-5.6 Sol & \scorepair{95.9}{86.7} & \scorepair{94.4}{83.3} & \scorepair{87.0}{50.0} & \scorepair{70.5}{20.0} & \scorepair{65.6}{13.3} \\
\bottomrule
\end{tabular*}
\end{table}

\begin{table}[!htbp]
\centering\small
\setlength{\tabcolsep}{4pt}
\caption{Local-goal completion / task success (LGC / TS, \%) by task family, pooling the five workload bands. Each cell gives equal weight to 50 executions, or 75 for each FieldOps scenario. Model labels abbreviate the full names in the experimental setup.}
\label{tab:dependency-families}
\begin{tabularx}{\linewidth}{@{}Yrrrrr@{}}
\toprule
Family & Grok & Qwen & DeepSeek & Gemini & GPT \\
\midrule
\multicolumn{6}{@{}l}{\textit{SocialOps}} \\
Overlapping conversions & \scorepair{99.1}{98.0} & \scorepair{97.0}{92.0} & \scorepair{96.8}{92.0} & \scorepair{95.4}{92.0} & \scorepair{46.1}{22.0} \\
Cooldown reactivation & \scorepair{98.0}{94.0} & \scorepair{96.7}{90.0} & \scorepair{96.3}{90.0} & \scorepair{69.8}{30.0} & \scorepair{40.6}{20.0} \\
Overlap + reactivation & \scorepair{96.6}{92.0} & \scorepair{90.7}{82.0} & \scorepair{97.1}{94.0} & \scorepair{79.5}{64.0} & \scorepair{48.4}{20.0} \\
\addlinespace[3pt]
\multicolumn{6}{@{}l}{\textit{LogisticsOps}} \\
Carrier disruption & \scorepair{97.5}{96.0} & \scorepair{42.8}{34.0} & \scorepair{58.3}{44.0} & \scorepair{52.0}{50.0} & \scorepair{51.5}{20.0} \\
Restricted handling & \scorepair{98.0}{94.0} & \scorepair{32.1}{18.0} & \scorepair{41.8}{22.0} & \scorepair{48.1}{44.0} & \scorepair{45.6}{4.0} \\
Specialist + disruption & \scorepair{97.2}{96.0} & \scorepair{32.3}{16.0} & \scorepair{45.3}{16.0} & \scorepair{46.8}{36.0} & \scorepair{48.5}{4.0} \\
\addlinespace[3pt]
\multicolumn{6}{@{}l}{\textit{ResearchOps}} \\
Identity compatibility & \scorepair{71.5}{54.0} & \scorepair{61.6}{44.0} & \scorepair{26.7}{6.0} & \scorepair{39.9}{22.0} & \scorepair{27.3}{12.0} \\
Study retraction & \scorepair{68.6}{58.0} & \scorepair{61.8}{40.0} & \scorepair{36.0}{14.0} & \scorepair{43.9}{24.0} & \scorepair{35.0}{16.0} \\
Ambiguity + withdrawal & \scorepair{70.6}{58.0} & \scorepair{58.6}{32.0} & \scorepair{27.2}{6.0} & \scorepair{46.0}{24.0} & \scorepair{25.7}{6.0} \\
\addlinespace[3pt]
\multicolumn{6}{@{}l}{\textit{DocOps}} \\
Expiring reservations & \scorepair{100.0}{100.0} & \scorepair{95.5}{82.0} & \scorepair{97.8}{76.0} & \scorepair{78.5}{74.0} & \scorepair{44.0}{14.0} \\
Independent amendments & \scorepair{100.0}{100.0} & \scorepair{96.7}{84.0} & \scorepair{100.0}{98.0} & \scorepair{80.8}{70.0} & \scorepair{68.2}{12.0} \\
Expiry + amendments & \scorepair{100.0}{100.0} & \scorepair{96.2}{86.0} & \scorepair{99.2}{84.0} & \scorepair{87.1}{82.0} & \scorepair{69.8}{26.0} \\
\addlinespace[3pt]
\multicolumn{6}{@{}l}{\textit{FactoryOps}} \\
Equipment windows & \scorepair{78.2}{48.0} & \scorepair{47.6}{16.0} & \scorepair{49.3}{18.0} & \scorepair{35.0}{12.0} & \scorepair{33.9}{2.0} \\
Selective rework & \scorepair{77.4}{42.0} & \scorepair{49.6}{18.0} & \scorepair{49.4}{14.0} & \scorepair{31.9}{12.0} & \scorepair{35.6}{4.0} \\
Constrained rework & \scorepair{78.6}{46.0} & \scorepair{43.2}{14.0} & \scorepair{45.0}{8.0} & \scorepair{37.3}{18.0} & \scorepair{34.7}{4.0} \\
\addlinespace[3pt]
\multicolumn{6}{@{}l}{\textit{FinanceOps}} \\
Heterogeneous integration & \scorepair{74.3}{40.0} & \scorepair{79.4}{32.0} & \scorepair{84.8}{26.0} & \scorepair{73.8}{34.0} & \scorepair{79.0}{6.0} \\
Receivables reconciliation & \scorepair{48.5}{8.0} & \scorepair{74.8}{0.0} & \scorepair{84.6}{6.0} & \scorepair{58.2}{2.0} & \scorepair{97.0}{28.0} \\
Release maintenance & \scorepair{75.5}{60.0} & \scorepair{99.9}{78.0} & \scorepair{99.7}{66.0} & \scorepair{87.4}{80.0} & \scorepair{95.0}{18.0} \\
\addlinespace[3pt]
\multicolumn{6}{@{}l}{\textit{FieldOps}} \\
Mesh convergence & \scorepair{93.2}{77.3} & \scorepair{89.1}{73.3} & \scorepair{96.6}{74.7} & \scorepair{50.6}{44.0} & \scorepair{87.5}{58.7} \\
Mesh supply shocks & \scorepair{90.5}{72.0} & \scorepair{84.0}{60.0} & \scorepair{89.0}{65.3} & \scorepair{55.0}{44.0} & \scorepair{77.8}{42.7} \\
\bottomrule
\end{tabularx}
\end{table}

\subsection{Execution Costs}
\label{app:execution-cost}
Table~\ref{tab:execution-cost} reports cumulative costs, counting shared continuation prefixes once.
\begin{table}[t]
\centering\small
\caption{Cumulative costs for 1,050 executions per model and reference-relative costs for the successful subset ($n$). $U$ counts charged attempts and $R$ reference calls. Compactions count adopted history replacements; utilization is median $100U/B^{\mathrm{run}}$ using actual allowances. P90 is the 90th percentile.}
\label{tab:execution-cost}
\setlength{\tabcolsep}{3pt}
\begin{tabular*}{\linewidth}{@{\extracolsep{\fill}}lrrrrrr@{}}
\toprule
 & \multicolumn{3}{c}{All executions} & \multicolumn{3}{c}{Successful executions} \\
\cmidrule(lr){2-4}\cmidrule(lr){5-7}
Model & Attempts & Compactions & Util. (\%) & $n$ & Median $U/R$ & P90 $U/R$ \\
\midrule
Grok 4.6 & 1,224,271 & 4,182 & 76.7 & 754 & $1.10\times$ & $1.26\times$ \\
Qwen3.8-Flash & 952,604 & 1,818 & 73.6 & 529 & $1.08\times$ & $1.20\times$ \\
DeepSeek V4.1 Flash & 1,162,694 & 3,379 & 75.9 & 495 & $1.12\times$ & $1.27\times$ \\
Gemini 3.8 Flash & 583,490 & 1,578 & 70.7 & 451 & $1.11\times$ & $1.30\times$ \\
GPT-5.6 Sol & 1,134,881 & 2,898 & 82.3 & 195 & $1.11\times$ & $1.35\times$ \\
\bottomrule
\end{tabular*}
\end{table}

\subsection{Task-Sampling Sensitivity}
\label{app:task-sampling}

The Gemini 3.8 Flash pool at 128K contains 167 heterogeneous-integration, 167 receivables-reconciliation, and 166 release-maintenance FinanceOps tasks, with 50, one, and 162 successes, respectively.

For 1,000 random orders, each family is independently shuffled and the families interleaved in that order, keeping prefix family counts within one task of each other. Figure~\ref{fig:task-sampling} reports mean success and pointwise 5th--95th percentiles for prefixes of 50--500 tasks, using linear percentile interpolation. The ranges describe subsampling of fixed outcomes at fixed family composition; prefixes share tasks across sample sizes.

\FloatBarrier
\section{Supporting Trace Analyses}
\label{app:unified-process}
\subsection{Progress and Stopping}
\label{app:sustained-progress}
\label{app:joint-analysis}
Absolute progress uses all 5,250 executions, with 30 per environment--model--band cell and equal weight per execution. Figure~\ref{fig:native-output-counts} retains the domain-specific units of Appendix~\ref{app:metrics}; different workload bands contain different task instances. Incomplete goals include blocked and unattempted work.

\begin{figure}[!htbp]
\centering
\includegraphics[width=\linewidth]{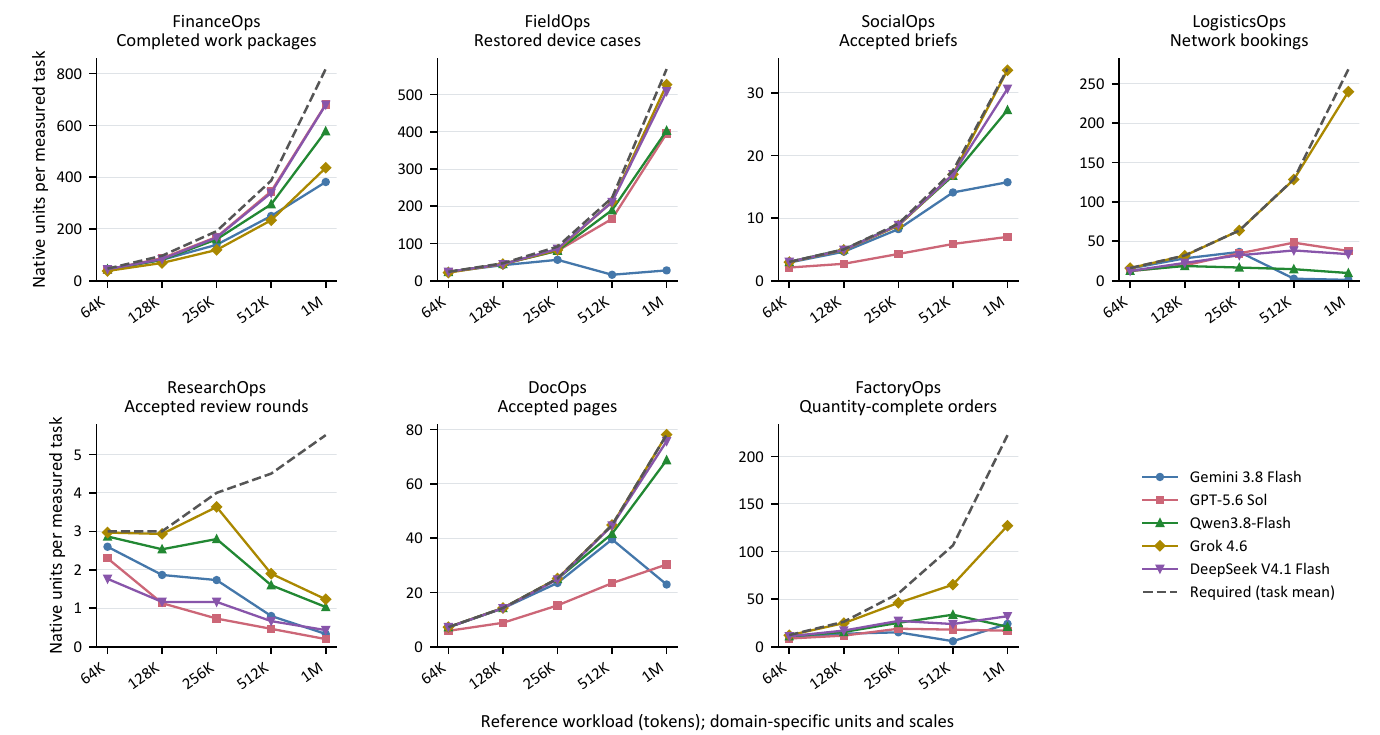}
\caption{Mean completed native units (solid) and required units (dashed), by model and environment. All 30 assigned executions contribute to each cell; required counts are identical across models. Panels retain domain-specific scales.}
\label{fig:native-output-counts}
\end{figure}

Stopping analyses examine agent-submitted final answers that fail business acceptance. Completion and unused allowance are jointly measured for 2,445 of 2,558 such submissions. Unused allowance is $(B-A_{\rm used})/B$, using charged attempts and the execution's recorded budget. Table~\ref{tab:joint-stopping-event} gives the denominators and events satisfying both LGC below 25\% and unused allowance above 50\%. Its all-assignment rate counts observed events among 1,050 executions per band; missing measurements are not imputed. With unused-allowance thresholds of 40\% and 60\%, failed-final event rates are both 2.4\% at 64K and respectively 58.9\% and 47.5\% at 1M; both alternatives increase across bands.
\begin{table}[!htbp]
\centering\small
\caption{Failed finals with LGC below 25\% and unused allowance above 50\%. Measured/all finals reports joint measurement coverage; all-assignment rates use 1,050 executions per band.}
\label{tab:joint-stopping-event}
\begin{tabular}{lrrrr}
\toprule
Workload & Measured/all finals & Events & Measured rate (\%) & All assigned (\%) \\
\midrule
64K & 206/208 & 5 & 2.4 & 0.5 \\
128K & 406/412 & 21 & 5.2 & 2.0 \\
256K & 518/543 & 79 & 15.3 & 7.5 \\
512K & 609/653 & 191 & 31.4 & 18.2 \\
1M & 706/742 & 381 & 54.0 & 36.3 \\
\bottomrule
\end{tabular}
\end{table}

\subsection{Local Execution and Global Delivery}
\label{app:skill-composition}
For Figure~\ref{fig:local-delivery}, full local completion requires equality of native integer completed and required counts. Conditional delivery at threshold $q$ selects executions satisfying $100g_i\geq qG_i$ and gives each eligible execution equal weight. Table~\ref{tab:threshold-delivery} reports the resulting task-success rates; increasing $q$ retains all successes while changing the eligible model and environment mix. Eleven executions complete all local goals but fail TS: two FinanceOps executions at 64K and nine FieldOps executions at 1M. These local/global distinctions follow the environment-specific definitions in Appendix~\ref{app:metrics}.
\begin{table}[!htbp]
\centering\small
\caption{Task success among executions with terminal LGC at least $q$, pooling seven environments and five models. The accompanying data retain the successful and eligible counts for all thresholds plotted in Figure~\ref{fig:local-delivery}.}
\label{tab:threshold-delivery}
\begin{tabular}{lrrrr}
\toprule
 & \multicolumn{4}{c}{Task success (\%)} \\
\cmidrule(l){2-5}
Workload & $q=75\%$ & $80\%$ & $90\%$ & $100\%$ \\
\midrule
64K & 88.6 & 92.1 & 95.3 & 99.7 \\
128K & 79.7 & 85.8 & 93.3 & 100.0 \\
256K & 75.0 & 80.2 & 87.3 & 100.0 \\
512K & 74.5 & 75.9 & 83.1 & 100.0 \\
1M & 62.2 & 65.7 & 71.6 & 96.3 \\
\bottomrule
\end{tabular}
\end{table}

To test whether successful script outputs satisfy downstream business checks, we match public returns to consumer arguments in later model decisions. Each consumer is counted once; when several earlier outputs match, all remain compatible sources. Same-decision calls are excluded because their arguments precede feedback. SocialOps matching checks post identity, historical-publication lists with duplicates preserved, and audience-load values with numeric types retained. Its 24,713 matched attempts span 730 executions: 21,326 are accepted and 3,387 rejected. Table~\ref{tab:skill-input-rejections} reports public reasons and rejection fractions computed within each execution and averaged equally across eligible executions. Matches identify consistent fields; source attribution and historical-input validity remain separate questions. FactoryOps accepts 1,494 of 1,495 matched finalizations across 111 executions, with one rejection for missing mandatory observations.
\begin{table}[!htbp]
\centering\small
\caption{SocialOps consumer rejection after matching a successful script output. Attempt counts and task-averaged rejection fractions have different denominators.}
\label{tab:skill-input-rejections}
\begin{tabular}{lr}
\toprule
Public rejection reason & Attempts \\
\midrule
Required historical source identities & 1,796 \\
Current decayed audience loads & 1,222 \\
Due-post time or status & 366 \\
Schema / post lookup & 2 / 1 \\
\midrule
Total rejected / matched attempts & 3,387 / 24,713 \\
\bottomrule
\end{tabular}
\par\smallskip
\begin{tabular}{lrrrrr}
\toprule
Workload & 64K & 128K & 256K & 512K & 1M \\
\midrule
Eligible executions & 142 & 147 & 148 & 147 & 146 \\
Mean rejection rate (\%) & 3.4 & 5.6 & 10.1 & 12.4 & 16.5 \\
\bottomrule
\end{tabular}
\end{table}

\subsection{Recovery}
\label{app:goal-followup}
Recovery follows the goal cycles defined in Appendix~\ref{sec:recovery}. Of 42,671 first-failure records, 35,265 cycles in 3,471 executions have an identifiable goal, public business-failure feedback, sufficient state/history, and measurable remaining allowance. Offline content findings and schema-only refusals are excluded. Eligibility is fixed at the first business failure and includes goals never checked again. A retry is a subsequent check-bearing attempt on that cycle, excluding the initial failure, reads, unrelated work, and edits without verification. Recovery within $k$ requires a pass by check $k$; all $k=3,\ldots,7$ use the same cycles. Fractions are averaged within executions, then equally across eligible executions in each model--band cell (Table~\ref{tab:recovery-opportunity-coverage}). At $k=7$, no-pass goals with fewer than seven checks account for 10.4\% at 64K and 24.7\% at 1M, versus 1.7\% and 2.3\% reaching seven checks without a pass.

\begin{table}[!htbp]
\centering\small
\caption{Eligible executions / first-failure cycles for the recovery curves. Every retry cutoff uses these same populations; their environment composition varies across cells.}
\label{tab:recovery-opportunity-coverage}
\begin{tabular}{lrrrrr}
\toprule
Model & 64K & 128K & 256K & 512K & 1M \\
\midrule
Grok & 68/131 & 91/375 & 107/853 & 118/1443 & 129/3484 \\
Qwen & 122/335 & 150/709 & 160/1048 & 162/2097 & 166/2251 \\
DeepSeek & 135/335 & 140/897 & 152/1138 & 158/1348 & 170/3432 \\
Gemini & 107/333 & 132/793 & 143/891 & 97/426 & 92/1174 \\
GPT & 159/602 & 174/1283 & 179/2069 & 180/3584 & 180/4234 \\
\bottomrule
\end{tabular}
\end{table}

Table~\ref{tab:recovery-outcomes} summarizes what happens to failed goals over the remaining execution: a later check passes, checks continue without a pass, or the goal receives no further check. These rates include all observed subsequent attempts. The retry-limited curves in Figure~\ref{fig:recovery-combined}a instead count passes within three to seven additional checks.
\begin{table}[!htbp]
\centering\small
\caption{Observed outcomes after a public business failure. Percentages are averaged within each execution and then across eligible executions in each workload band. All subsequent checks up to the observed endpoint are included.}
\label{tab:recovery-outcomes}
\begin{tabular}{lrrrrr}
\toprule
Workload & Executions & Failed goals & Recovered (\%) & Rechecked, no pass (\%) & No recheck (\%) \\
\midrule
64K & 591 & 1,736 & 88.7 & 5.8 & 5.4 \\
128K & 687 & 4,057 & 82.6 & 11.0 & 6.5 \\
256K & 741 & 5,999 & 80.5 & 12.7 & 6.8 \\
512K & 715 & 8,898 & 79.7 & 14.6 & 5.8 \\
1M & 737 & 14,575 & 74.1 & 15.8 & 10.1 \\
\bottomrule
\end{tabular}
\end{table}

\paragraph{Observed budget prefixes.}
\label{app:budget-prefix}
Figure~\ref{fig:recovery-combined}b--d evaluates fixed prefixes of 3,374 recorded executions containing 30,452 cycles first failing by the reference allowance $R$. The five bands contribute 558, 659, 722, 704, and 731 executions. At each reference threshold $B_\lambda=R+\lceil\lambda A\rceil$, $\lambda\in\{0,1/8,1/4,3/8,1/2\}$, the prefix ends at $\min(B_\lambda,U_i)$, where $U_i$ is the observed terminal charged-attempt count; earlier terminal states are retained. Charges include individual attempted actions within batches and charged refusals, excluding exempt declarations. Native completion checks score the state at that prefix; recovery requires a subsequent pass within the same cycle and before the cutoff. All three metrics use the same execution cohort. LGC covers all required local goals, and recovery covers eligible failed cycles.

\subsection{Dependency Visibility across Compaction}
\label{app:required-dependency-compaction}
Both local comparison groups act after an adopted summary. Immediately before the first attributable business attempt, required artifact roles are identified from the task contract and preceding public state, independently of whether submitted inputs are correct. The retained group has every identified candidate available as a complete structured tool return in the consuming request. The removed group has at least one required role whose candidates were all removed by an adopted summary and remain absent at consumption. Visibility requires the artifact's identity and payload together in a complete structured return; a complete reread restores visibility. Mixed alternatives, unresolved requests, unknown visibility, and unattempted targets are excluded. Environment-specific artifact roles define the scope of the comparison; coverage of resource, numeric, and timing constraints is partial.

The paired population is 1,518 executions from 599 base tasks, with 2,610 execution--operation--cycle-kind cells. First-check acceptance measures the initial outcome; ever accepted records whether any check passes before cycle end. Additional checks count subsequent checks until acceptance or cycle cutoff, including unresolved goals and assigning zero to first-pass goals. Targets are averaged within group and cell, cells within executions, executions within environments, and the seven environments equally. Table~\ref{tab:required-dependency-compaction} reports acceptance rates and additional checks for both groups in each environment. Pairs share the same execution and operation; goals can differ in difficulty, timing, and state.

The task-level comparison separately counts summaries adopted in subsequent model or compaction requests, deduplicating continuation prefixes. Among all 5,250 executions, 95 of 175 environment--workload--model strata contain both zero- and nonzero-compaction groups, contributing 1,116 and 1,734 executions. Table~\ref{tab:compaction-task-outcomes} reports completion, task success, and tool calls with equal weights over these shared strata. Compaction exposure depends on execution duration: agents can stop before the first compaction, while longer executions provide more opportunities for compaction.
\begin{table}[!htbp]
\centering\small
\caption{Business checks with required inputs retained or removed from the active context. Each environment uses executions with observations in both groups; the overall row weights the seven environments equally. Acceptance rates are percentages.}
\label{tab:required-dependency-compaction}
\setlength{\tabcolsep}{4pt}
\begin{tabular*}{\linewidth}{@{\extracolsep{\fill}}lrrrrrrr@{}}
\toprule
 & & \multicolumn{2}{c}{First pass (\%)} & \multicolumn{2}{c}{Ever accepted (\%)} & \multicolumn{2}{c}{Additional checks} \\
\cmidrule(lr){3-4}\cmidrule(lr){5-6}\cmidrule(l){7-8}
Environment & Executions & Retained & Removed & Retained & Removed & Retained & Removed \\
\midrule
FinanceOps & 342 & 50.7 & 47.1 & 51.1 & 51.1 & 0.16 & 0.28 \\
FieldOps & 302 & 96.2 & 94.3 & 99.3 & 99.2 & 0.05 & 0.06 \\
SocialOps & 137 & 95.4 & 86.8 & 100.0 & 100.0 & 0.08 & 0.22 \\
LogisticsOps & 159 & 77.1 & 67.0 & 97.3 & 91.7 & 0.49 & 1.05 \\
ResearchOps & 120 & 99.7 & 99.6 & 100.0 & 100.0 & 0.00 & 0.00 \\
DocOps & 277 & 87.3 & 67.6 & 99.0 & 97.6 & 0.27 & 0.63 \\
FactoryOps & 181 & 85.2 & 78.2 & 95.6 & 89.1 & 0.14 & 0.16 \\
\midrule
Overall & 1,518 & 84.5 & 77.2 & 91.8 & 89.8 & 0.17 & 0.34 \\
\bottomrule
\end{tabular*}
\end{table}
\begin{table}[!htbp]
\centering\small
\caption{Observed task outcomes with and without compaction. Both groups receive equal weights over the same 95 environment--workload--model strata.}
\label{tab:compaction-task-outcomes}
\begin{tabular}{lrrrr}
\toprule
Compaction & Executions & LGC (\%) & TS (\%) & Mean tool calls \\
\midrule
None & 1,116 & 37.0 & 23.4 & 303 \\
At least one & 1,734 & 71.3 & 41.1 & 1,159 \\
\bottomrule
\end{tabular}
\end{table}

\FloatBarrier
\subsection{Sustained Execution and Harness Sensitivity}
\label{app:gpt-harness}

\paragraph{Gemini's strong short-task performance gives way to early termination.}
Gemini 3.8 Flash achieves 92.3\% LGC and 78.6\% TS at 64K, falling to 21.3\% and 8.1\% at 1M. Its unsuccessful final answers increasingly occur with little completed work and substantial unused allowance. As shown in Table~\ref{tab:model-stopping}, median unused allowance rises from 18.1\% at 64K to 96.1\% at 1M. Among the measured 1M failed finals, 83.0\% complete less than one-quarter of local goals while retaining more than half the allowance. Gemini can complete the shorter workflows, but its longer executions frequently terminate before making substantial progress.

\paragraph{Unfinished work increasingly remains when GPT stops.}
GPT-5.6 Sol completes 195 of 1,050 tasks in the main evaluation. Of its 855 unsuccessful executions, 746 end with an agent-submitted final answer. The median LGC at failed finals falls from 66.7\% at 64K to 17.1\% at 1M, while median unused allowance rises from 8.7\% to 53.9\%. At 1M, 48.1\% of the measured failed finals meet the same low-progress, high-allowance condition. Early termination is therefore shared by these two models, with a larger observed rate for Gemini at the longest workload.
\begin{table}[htbp]
\centering\small
\caption{Stopping behavior at unsuccessful agent-submitted final answers. LGC and unused allowance are medians. The last column gives the share with LGC below 25\% and unused allowance above 50\%. $n$ counts executions with both measurements; other values are percentages.}
\label{tab:model-stopping}
\setlength{\tabcolsep}{3pt}
\begin{tabular*}{\linewidth}{@{\extracolsep{\fill}}lrrrr@{}}
\toprule
Workload & $n$ & LGC & Unused allowance & Low progress, high allowance \\
\midrule
\multicolumn{5}{@{}l}{\textit{Gemini 3.8 Flash}} \\
64K & 41 & 66.7 & 18.1 & 0.0 \\
128K & 82 & 66.1 & 24.3 & 15.9 \\
256K & 112 & 25.0 & 44.7 & 36.6 \\
512K & 153 & 3.2 & 90.8 & 67.3 \\
1M & 188 & 1.2 & 96.1 & 83.0 \\
\midrule
\multicolumn{5}{@{}l}{\textit{GPT-5.6 Sol}} \\
64K & 70 & 66.7 & 8.7 & 0.0 \\
128K & 123 & 50.0 & 12.0 & 1.6 \\
256K & 166 & 45.6 & 14.5 & 3.0 \\
512K & 181 & 34.2 & 25.9 & 16.0 \\
1M & 183 & 17.1 & 53.9 & 48.1 \\
\bottomrule
\end{tabular*}
\end{table}

\paragraph{Early stopping can occur before any context compaction.}
In a 1M DocOps task, GPT completes 11 of 81 required pages after using 172 of its 1,715 permitted tool attempts. It submits a final answer, receives the neutral confirmation prompt, and submits another final answer without further tool use. No compaction has occurred, and 90.0\% of the allowance remains. This trajectory shows that stopping decisions can leave an ongoing workflow unfinished even before compaction or budget exhaustion.

\paragraph{The same tasks yield substantially different outcomes under Codex.}
We examine a supplementary 50-task FinanceOps experiment using GPT-5.6 Sol with medium reasoning in Codex. The selected tasks span all five workload bands and retain their reference-derived tool allowances. Codex accesses the environment through native MCP calls to direct business operations; these operations forward to the FinanceOps runtime and its acceptance checks. Shell access, web access, and multi-agent execution are disabled. Interrupted runs resume from their saved business states and remaining allowances.

As shown in Table~\ref{tab:gpt-harness}, task success rises from 18.0\% under Harbor + Inspect to 84.0\% under the Codex configuration. Unsuccessful final answers decrease from 38 to eight, and no Codex task exhausts its tool allowance. The comparison changes the tool interface, provider route, and context-management policy together: Codex uses direct native MCP operations and its own automatic compaction, while the main evaluation uses the skill-facing interface and the shared summary-based policy. GPT's task success varies markedly across these agent configurations.
\begin{table}[htbp]
\centering\small
\caption{GPT-5.6 Sol on the same 50 FinanceOps tasks, with ten tasks per workload band. Both configurations request medium reasoning. TS is a percentage; other entries are task counts.}
\label{tab:gpt-harness}
\setlength{\tabcolsep}{3pt}
\begin{tabular*}{\linewidth}{@{\extracolsep{\fill}}lrrrr@{}}
\toprule
Configuration & Success & Final failure & $B$ exhausted & TS \\
\midrule
Harbor + Inspect & 9 & 38 & 3 & 18.0 \\
Codex + direct MCP & 42 & 8 & 0 & 84.0 \\
\bottomrule
\end{tabular*}
\end{table}

\FloatBarrier
\subsection{Local Completion and Global Delivery}
\label{app:model-delivery}

\paragraph{DeepSeek sustains substantial local progress without consistently completing delivery.}
DeepSeek V4.1 Flash retains 56.1\% LGC at 1M, compared with 48.9\% for Qwen3.8-Flash and 21.3\% for Gemini 3.8 Flash. Its outcomes nevertheless expose a substantial gap between local progress and whole-task acceptance. As shown in Table~\ref{tab:deepseek-delivery}, 1M LGC remains 97.2\% in DocOps, 84.7\% in FieldOps, and 85.0\% in FinanceOps, while TS reaches 66.7\%, 30.0\%, and 6.7\%, respectively.
\begin{table}[htbp]
\centering\small
\caption{DeepSeek V4.1 Flash at 1M in three environments. Each row includes 30 tasks. LGC and TS are percentages; the last column counts unsuccessful tasks with at least 75\% LGC.}
\label{tab:deepseek-delivery}
\setlength{\tabcolsep}{3pt}
\begin{tabular*}{\linewidth}{@{\extracolsep{\fill}}lrrr@{}}
\toprule
Environment & LGC & TS & High-completion failures \\
\midrule
FinanceOps & 85.0 & 6.7 & 21 \\
FieldOps & 84.7 & 30.0 & 14 \\
DocOps & 97.2 & 66.7 & 8 \\
\bottomrule
\end{tabular*}
\end{table}

\paragraph{The delivery gap occurs within individual executions.}
Among the 30 FinanceOps tasks at 1M, 21 finish with at least 75\% of local goals satisfied but fail whole-task acceptance. FieldOps and DocOps contain another 14 and eight such executions, respectively. Substantial local progress remains unfinished at the workflow level.

\FloatBarrier
\subsection{Task Configuration and Model Strengths}
\label{app:grok-selection}

\paragraph{Grok demonstrates strong sustained execution.}
Grok 4.6 achieves the highest overall LGC and TS, reaching 86.0\% and 71.8\%, respectively; the next-highest overall TS is 50.4\%. Its TS reaches 100.0\% in DocOps, 95.3\% in LogisticsOps, and 94.7\% in SocialOps. The same LogisticsOps tasks yield at most 43.3\% TS among the other four models. These differences show substantial discrimination even in a domain where Grok approaches saturation. A successful 1M LogisticsOps trajectory completes all 256 bookings after 3,571 tool attempts, including 170 tool errors and only one script invocation. Its accepted terminal business state demonstrates sustained execution through errors and subsequent correction.

\paragraph{High scores characterize the selected workflow configurations.}
The evaluated tasks instantiate a configurable environment at particular business scales, dependency patterns, resource conditions, and update settings. Selection uses reference feasibility and workload, as described in Appendix~\ref{app:generation-rules}; increasing reference workload extends the work performed under these chosen rules. In DocOps, the selected range grows from 6--10 pages at 64K to 52--105 pages at 1M, with configured depth reaching 35. These dependencies support a repeatable procedure: obtain current parent artifacts, compute a contract with the supplied script, submit it for review, and complete publication and tracker synchronization. Scripts handle deterministic contract calculations, while the agent selects inputs, maintains current bindings, and carries the workflow through delivery and updates. Grok's high scores show that it can sustain this procedure across the tested scales. Reference workload and the difficulty of each successive decision are therefore distinct properties of the selected tasks.

\paragraph{Other configurations reveal substantial remaining difficulty.}
As shown in Table~\ref{tab:detailed-results}, Grok's 1M LGC falls to 22.3\% in ResearchOps and 56.3\% in FactoryOps, with zero TS in both. These workflows require cumulative evidence integration or coordination of production state. Variation also appears within a single environment: Grok's FinanceOps TS is 40.0\% for heterogeneous integration, 8.0\% for receivables reconciliation, and 60.0\% for release maintenance, as shown in Table~\ref{tab:dependency-families}. Thus, the evaluation contains both profiles that Grok handles reliably and profiles that remain difficult. The near-perfect scores locate the former within the tested configuration range. The generators expose scale, dependency, resource, and update parameters separately, allowing subsequent evaluations to examine additional configurations beyond this selection.

\paragraph{Qwen's relative strengths vary with workflow and scale.}
Qwen3.8-Flash and DeepSeek have nearly identical overall LGC, at 72.2\% and 72.1\%, but Qwen achieves higher overall TS, at 50.4\% versus 47.1\%. At 1M, DeepSeek leads on both metrics, reaching 56.1\% LGC and 26.2\% TS compared with Qwen's 48.9\% and 22.9\%. Task-family results further distinguish their strengths. Within FinanceOps, Qwen achieves 99.9\% LGC and 78.0\% TS on release maintenance, 79.4\% and 32.0\% on heterogeneous integration, and 74.8\% and zero TS on receivables reconciliation. Across ResearchOps, Qwen achieves 38.7\% TS, whereas DeepSeek reaches 8.7\%. These results show why model comparisons need both workload and workflow breakdowns: an overall advantage can reverse at larger scales or under different business requirements.

\clearpage
\section{Continual Task Generation and Future Testbed Uses}
\label{app:generation-testbed}

\subsection{Instance Generation and Horizon Extension}
\label{app:generation-extension}

\paragraph{A renewable source of tasks.}
\bench{} specifies task families through business rules, construction parameters, and seeds. Seeds sample business objects, attributes, and connections within a configuration. Construction parameters vary the amount of required work, dependency structure, and resource or update conditions. Together, they support replenishing evaluation sets and testing transfer across configurations. Task diversity is characterized by variation in these business properties and the decisions they require.

\paragraph{Extending required business work.}
Task composition provides a route toward indefinitely extensible workflows: add required activities, connect them to earlier outputs, and extend the associated resources and completion requirements. Sequential dependencies lengthen chains of obligations, joins combine multiple results, and long-range reuse connects later work to earlier evidence. Shared resources and updates introduce dependencies through changing business state. Public reference execution checks that the enlarged workflow remains feasible. The released builders use bounded construction profiles; further scale extension couples broader profiles with resource, update, and reference-execution validation. This approach makes business obligations and their connections the basis of horizon growth.

\paragraph{Business horizon and context length.}
Longer workflows generate cumulative interaction histories that can be processed through repeated compaction under a fixed input capacity. Business state and artifacts remain accessible through public tools, allowing agents to recover earlier evidence and continue execution. Reference workload measures the interactions of the public reference execution, while active-context length measures the information presented at a particular step. Recording both alongside business work counts links execution demands to information retention and recovery.

\subsection{A Testbed for Improving Long-Horizon Agents}
\label{app:testbed-outlook}

\paragraph{Context and dependency management.}
One future use is to compare summarization, retrieval, and dependency-aware memory on identical generated tasks, with the same environment, business checks, tools, and execution allowance. First-attempt correctness, recovery from stale or missing inputs, final completion, and evidence-retrieval cost capture both the reliability of continuation and the effort required to maintain it.

\paragraph{Skills and execution frameworks.}
Another direction is to evaluate changes to skill documents, supplied scripts, and agent controllers \citep{liu2026skillrevise,liu2026skillscriptbench,liu2026rethinking}. Fixing each candidate before testing on held-out seeds and configurations allows evaluation of transfer to new tasks and larger workloads. State-based business checks provide common outcome measures across systems that choose different action sequences. Process measurements connect changes in scheduling, handoffs, and recovery to sustained progress and complete delivery.

\paragraph{Iterative and recursive self-improvement.}
Continual generation also offers a testbed for repeated improvement cycles. A prospective protocol alternates revision of skills, memory policies, or controllers on development trajectories with evaluation of the resulting agent on fresh tasks. Held-out configurations test transfer to new dependency and update patterns, while improvement and execution costs quantify the resources used across cycles. Recursive self-improvement research can extend this protocol to evaluate changes to the improvement procedure itself, alongside changes to the task-solving agent.

\end{document}